%% file: tmlr.tex
\documentclass[10pt]{article} 
\usepackage[preprint]{tmlr}

\input{math_commands.tex}

\usepackage{hyperref}
\usepackage{url}

\usepackage{amsmath,amssymb,amsfonts}
\usepackage{graphicx}
\usepackage{adjustbox}
\usepackage{tablefootnote}
\usepackage{colortbl}
\usepackage{booktabs}
\usepackage{arydshln}

\title{Answering Unseen Questions With Smaller Language\\Models Using Rationale Generation and Dense Retrieval}

\author{\name Tim Hartill \email thar011@aucklanduni.ac.nz \\
      \addr School of Computer Science\\
      University of Auckland
      \AND
      \name Diana Benavides-Prado \email d.benavides-prado@auckland.ac.nz \\
      University of Auckland
      \AND
      \name Michael Witbrock \email m.witbrock@auckland.ac.nz\\
      \addr School of Computer Science\\
      University of Auckland
      \AND
      \name Patricia J. Riddle \email p.riddle@auckland.ac.nz\\
      \addr School of Computer Science\\
      University of Auckland}

\def\month{MM}  
\def\year{YYYY} 
\def\openreview{\url{https://openreview.net/forum?id=XXXX}} 

\begin{document}

\maketitle

\begin{abstract}

When provided with sufficient explanatory context, smaller Language Models have been shown to exhibit strong reasoning ability on challenging short-answer question-answering tasks where the questions are unseen in training. We evaluate two methods for further improvement in this setting. Both methods focus on combining rationales generated by a larger Language Model with longer contexts created from a multi-hop dense retrieval system. The first method (\textit{RR}) involves training a Rationale Ranking model to score both generated rationales and retrieved contexts with respect to relevance and truthfulness. We then use the scores to derive combined contexts from both knowledge sources using a number of combinatory strategies. For the second method (\textit{RATD}) we utilise retrieval-augmented training datasets developed by \citet{Hartill2023-pf} to train a smaller Reasoning model such that it becomes proficient at utilising relevant information from longer text sequences that may be only partially evidential and frequently contain many irrelevant sentences. We find that both methods significantly improve results. Our single best Reasoning model materially improves upon strong comparable prior baselines for unseen evaluation datasets (StrategyQA 58.9 $\rightarrow$ 61.7 acc., CommonsenseQA 63.6 $\rightarrow$ 72.7 acc., ARC-DA 31.6 $\rightarrow$ 52.1 F1, IIRC 25.5 $\rightarrow$ 27.3 F1) and a version utilising our prior knowledge of each type of question in selecting a context combination strategy does even better. Our proposed models also generally outperform direct prompts against much larger models (BLOOM 175B and StableVicuna 13B) in both few-shot chain-of-thought and standard few-shot settings.
 
\end{abstract}

\section{Introduction}

\begin{figure*}[h]
\begin{center}
\includegraphics[width=\textwidth]{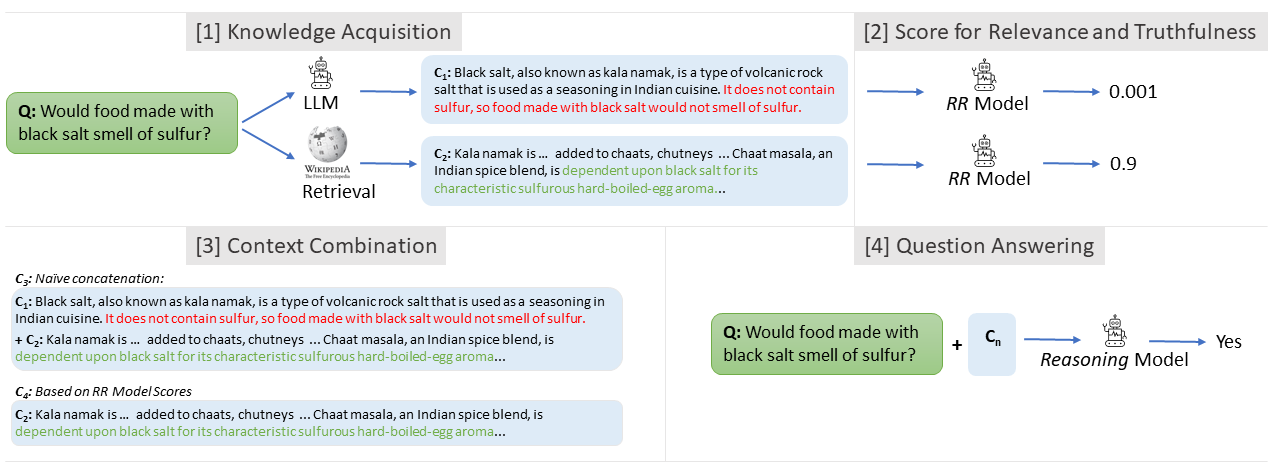}
\caption{Overview of our approach. Given an unseen question \textbf{Q}: [1] we acquire explanatory contexts, \textbf{C\textsubscript{1}} and \textbf{C\textsubscript{2}}, from two knowledge sources. [2] We score the acquired contexts for relevance and truthfulness using a Rationale Ranking (\textit{RR}) model that we train on diverse relevant/irrelevant samples that make both truthful and false assertions. [3] We evaluate and select methods for combining or filtering \textbf{C\textsubscript{1}} and \textbf{C\textsubscript{2}}. [4] We evaluate the performance of different contexts (\textbf{C\textsubscript{n}}) on a set of Reasoning Models that are trained on different mixtures of training datasets, including a mixture containing \textit{RATD} datasets \citep{Hartill2023-pf} and a mixture without these. In the diagram, \textcolor{red}{red} denotes false information and {\color[HTML]{10E689}green} highlights relevant and truthful evidence.}
\label{ovw_fig}
\end{center}
\end{figure*} 

\textit{``It was soon realized that the problem of systematically acquiring information from the environment was much less tractable than the mental activities the information was intended to serve'' - \citet{Moravec1988-yw} }

Moravec's paradox is the observation that problems such as developing an ability to reason, that might have been assumed to be one of the most difficult challenges in artificial intelligence has been easier to resolve than the challenge of acquiring more basic knowledge such as sensory information. It is motivating to consider this in the context of recent advances in using both large Language Models (LLMs) and retrieval against large textual corpora for information acquisition in the question-answering domain. 

We focus on methods to improve the performance of a smaller Language Model\footnote{Generative Transformers with 400 million to 1 billion parameters} (i.e. Reasoning Model) which, given a question and an acquired explanatory context as input, is expected to reason to provide an answer. Our interest in smaller models for this task is because we are interested in evaluating the viability of reasoning systems that answer arbitrary questions in resource constrained situations where available compute resource is limited, and internet connectivity and so forth is assumed to be unavailable.

To acquire the explanatory context, we consider two knowledge sources separately and in combination; retrieval of an explanatory context from a corpus of English Wikipedia paragraphs and rationale\footnote{We use the term ``rationale'' to denote a free-text explanation \citep{Wiegreffe2021-go} of approximately one to three sentences that provides evidence to support a model prediction. We use the term to distinguish LLM generations of this form from the longer explanatory contexts produced from our retrieval system.} generation from LLMs. Retrieval has generally been a relatively resource-efficient activity but until recently even inference on LLMs has required considerable computational resources. Recent innovations such as those involving 8-bit matrix multiplication (INT8) \citep{Dettmers2022-te} enable the use of LLMs as frozen knowledge bases in constrained settings. For example inference on the 13 billion parameter StableVicuna model \citep{Stability-AI2023-mp} that we convert to INT8 and use in some experiments runs in approximately 18 GB of GPU RAM, well within the current capacity of large consumer GPU cards.

We choose retrieval from a reliable corpus and LLMs as our knowledge sources since we hypothesise that they may offer differing and complimentary characteristics. Studies such as \citet{Khattab2021-jf, Hartill2023-pf} have shown that multi-hop retrieval systems can be proficient at identifying the relevant $n$ documents necessary to answer $n$-hop factual questions where $n$ can be greater than two, e.g. those found in the Hover \citep{Jiang2020-on} or Musique \citep{Trivedi2022-mv} datasets (``The Rhine forms a border between Aschenbrödel's composer's country and another country where women got the vote when?''). However we are unaware of any corresponding studies on LLMs that demonstrate similar proficiency in generating sufficient information to answer such $n$-hop questions. Conversely, it has been shown that LLMs can be strong at answering commonsense questions without using external retrieval \citep{Lourie2021-kb}, but for such questions retrieval from large textual corpora offers limited benefit \citep{Piktus2021-lu, Hartill2023-pf}. 

We explore two methods of combining information from our knowledge sources. Our Rationale Ranking method (\textit{RR}) involves training a smaller Transformer to score both rationales and retrieved explanatory contexts with respect to relevance and truthfulness. We then evaluate a number of simple strategies to create combined contexts such as including either or both components that score over a threshold, or selecting the single top-scoring component. We focus on identifying combination methods that work best in the general case, i.e. are most likely to work well for an arbitrary unseen question for which we provide no means of predicting which combination method will work best. We find that we are able to identify such a method for each of our Reasoning Models and quantify the performance improvement (section \ref{rr_ratd_improvement quantification}). Our second method (\textit{RATD}) consists of training our Reasoning Model with retrieval-augmented datasets previously developed by \cite{Hartill2023-pf}. These datasets were originally developed to impart diverse reasoning strategies such as an ability to identify and weigh partially evidential facts in long, noisy contexts. Noting that where our rationales and retrieved contexts are combined, the resulting context is similar in length and form to the \textit{RATD} contexts, we find that training on them enables a single Reasoning Model to utilise our various context formats effectively, including the case where the context consists of the naïve concatenation of rationale and retrieved context that does not consider the \textit{RR} model scores.

In summary the major contributions of this paper are: (A) We propose \textit{RR}, a novel method that both selects context components by relevance, and filters components that may be false. (B) We apply the \textit{RATD} method developed by \citet{Hartill2023-pf} to facilitate reasoning over contexts that potentially combine information from multiple knowledge sources. (C) We demonstrate that both methods in isolation significantly improve reasoning performance in smaller Language Models from strong baselines in the same unseen setting (section \ref{rr_ratd_improvement quantification}). (D) We show that smaller Language Models can manifest comparable or stronger reasoning performance as a LLM when provided with the same knowledge to reason over that the LLM is capable of generating for itself (section \ref{gr_ratd_vs_llm}). (E) We illustrate the respective strengths and weaknesses of LLMs and multi-hop retrieval from a Wikipedia corpus as knowledge sources (section \ref{gr_ratd_vs_llm}). (F) We show that combining information from these sources significantly improves the overall average performance versus using a single source, often beyond what either knowledge source in isolation can deliver on individual datasets (section \ref{gr_ratd_vs_llm}).

\subsection{Related Work}
\label{related_work}

\textbf{Knowledge Augmentation from LLMs}. \citet{Bosselut2019-wn} proposed COMET, a GPT-based Model \citep{Radford2018-kr} trained on triples from the ATOMIC  \citep{Sap2019-tk} and ConceptNet \citep{Speer2017-ll} knowledge graphs such that it would generate potentially novel triple completions. \citet{Shwartz2020-sn} compare augmentation methods using COMET, ConceptNet and their self-talk method where the question-answering Language Model is self-queried to produce additional information pertinent to answering the question. \citet{Liu2022-bc} generate knowledge statements from GPT-3 \citep{Brown2020-rl} conditioned on the question and use the augmented samples in separate smaller Reasoning Models. Following the introduction of chain-of-thought (COT) prompting \citep{Wei2022-lz}, a number of recent papers use this prompting style to distill training sets of rationale-augmented samples from internet-accessable LLMs (GPT-3, Palm \citep{Chowdhery2022-yw}) which are then typically used to train much smaller models in task-specific finetuned settings e.g. \citep{Magister2023-uf, Li2023-ss, Hsieh2023-my, Wu2023-dg, Shridhar2023-qb} sometimes such that the label and the rationale are output to avoid the issue of having to generate a rationale from the LLM at test time. We note that our usage of LLM-generated rationales is rather different from these in that we assume a locally-accessable LLM  (with lower resource requirements) at test time and do not incorporate LLM-generated rationales in our Reasoning Model training.

\textbf{Retrieval from Textual Corpora}. For a comprehensive introduction to this wide field we suggest reviewing \citet{Lin2022-jq} and \citep{Mitra2018-mr}. In summary, TF-IDF \citep{Sparck_Jones1972-dm} has been used for many years to associate queries with documents using adjusted bag-of-word count vectors. This approach carries the advantage that fine-tuning for the target dataset is not required. \citet{Chen2017-gw} first used such sparse retrieval against Wikipedia in the context of open domain question-answering. In dense retrieval, query and corpus documents are embedded into the same vector space with similarity defined as the inner product between a query and a document vector. \citet{Karpukhin2020-wa} used dense retrieval to identify a single document sufficient for answering a single-hop question. \citet{Izacard2022-hu} reduce the need for target dataset finetuning by pretraining a dense retriever on self-supervised data. \citet{Xiong2021-ex} extend the dense retrieval approach to to retrieve two documents necessary to answer a complex two-hop question. \citet{Hartill2023-pf} extend this to enable retrieval of an arbitrary maximum number of documents (in practice $n \leq 4$). \citet{Wang2018-zz} introduced a Reranker Model that refines retrieved results. Baleen \citep{Khattab2021-jf} is a two-stage condenser system comprising a Reranker followed by an additional model that scores relevance of each sentence selected over multiple documents ($n \leq 4$) from the first stage. \citet{Hartill2023-pf} introduce an  Evidence Set Score into the second stage to quantify the sufficiency of the entire set of selected sentences for answering a query and call their resulting system the ``Iterator''. As noted, in this paper we use the Iterator with a Wikipedia corpus as described the following section.

\textbf{Multiple Knowledge Sources}. Retrieval has been successfully used as a method for querying  knowledge graphs by embedding the constituent triples as the document vectors in addition to, or instead of, standard text, e.g. \citet{Yu2022-be} augment commonsense questions with retrieved information from a commonsense-focused corpus consisting of information source from knowledge graphs, commonsense datasets and other textual sources. Perhaps most similar in spirit to our work \citet{Pan2023-sz} consider knowledge graphs, Wikipedia data, a dictionary, and others, as separate knowledge sources, each queried using dense retrieval. In contrast to our approach of considering various methods for combining information, they train a model to select the single most relevant source for augmenting each input sample. This is analogous to our ``Max Score'' method described in section \ref{sec:context_combination_strategies}. Like us they train a smaller Reasoning Model with disparate training and evaluation datasets, although unfortunately their evaluation datasets differ from ours. Also in a similar direction to our ``Max Score'' method, \citet{Si2023-kz} route a query to four expert LLMs and select the single most likely answer using a smaller classifier trained for that purpose.  In a finetuned setting, \citet{ijcai2022p383} also consider multiple knowledge sources. Here they use an entity linking method to query ConceptNet and sparse retrieval over a dictionary and a set of commonsense datasets. The results are always concatenated which is similar to our Naïve Concatenation method (section \ref{sec:context_combination_strategies}).

\textbf{Falsehood Detection}. Our \textit{RR} Model, trained to score for truthfulness and relevance over instances from disparate knowledge sources, can be seen as a novel extension to a Reranking approach. However it also shares an objective with methods aiming to detect falsehood in LLM generations. Generally these methods fall into three categories. The first are methods based on the intuition that higher token log probabilities correspond to better text along a particular dimension such as truthfulness \citep{Yuan2021-dm,Fu2023-ir}. The second are factuality detection methods that evaluate LLM-generated assertions as true if they can be supported by a external reference (e.g fact retrieval from a reliable corpus). Recent studies here include \citep{Min2023-fc, Chern2023-jw}. A third category, broadly called self-checking involves prompting a LLM such as ChatGPT or GPT-4 \citep{OpenAI2023-cv} to recognize their own errors \citep{Chern2023-jw}, or refine their own outputs \citep{Chen2023-lc,Madaan2023-rm}, without recourse to external tools. In this category but with a different approach, \citet{Manakul2023-db} score the consistency between a reference statement and several stochastically sampled versions of it that may be likely to diverge more if the reference is a hallucination.

\section{Method}
\label{sec:method_overview}

An overview of our approach is provided in Figure \ref{ovw_fig}. In following sections we describe how the two knowledge sources are implemented, how the \textit{RR} model is constructed, trained and initially evaluated, and how the Reasoning Models are trained. We describe our context combination methods further below in section \ref{sec:context_combination_strategies} so as to make clear the nomenclature we use in reporting experimental results.

A major assumption is that our system may be asked arbitrary questions from unknown distributions. Therefore we primarily consider our evaluations in the unseen rather than fine-tuned setting. The most relevant comparisons we have available to us are the baselines for StrategyQA \citep{Geva2021-sl}, CommonsenseQA \citep{Talmor2019-rm}, ARC-DA \citep{Bhakthavatsalam2021-fq}, IIRC \citep{Ferguson2020-hv} and Musique \citep{Trivedi2022-mv} established for smaller Language Models in unseen settings by \citet{Hartill2023-pf}. The datasets cover a diversity of question types requiring diverse reasoning strategies to answer, including commonsense and $n$-hop factual questions ($n \leq 4$) as discussed further in section \ref{sec:unseen_eval_datasets_desc}. Hence we adopt these datasets for evaluation and use the same definition as \citet{Hartill2023-pf} for ``seen-ness'' whereby an unseen evaluation sample is one from a dataset that is disjoint from any training dataset. In our case we extend this to our LLM generations, ensuring that all examples in few-shot prompts come from our training rather than evaluation datasets, or are manually created by us.

Aside from the baseline results, \citet{Hartill2023-pf} also provide their ``Iterator'' $n$-hop dense retrieval system (where $n \leq 4$). In a single-hop retrieval model, samples are processed as (1) Input $\langle q\rangle$ with an objective of minimizing distance to the vector representing $d\textsubscript{0}$ (hereafter denoted $\langle q\rangle \rightarrow d\textsubscript{0}$, where $q$ and $d\textsubscript{t}$ are the input question and the \textit{t-th} supporting document of $q$ to retrieve respectively). For a two hop system, the second hop is then (2) $\langle q, d\textsubscript{0}\rangle\rightarrow d\textsubscript{1}$. In the Iterator model this is extended up to $4$ hops i.e. $\langle q, d\textsubscript{0}, d\textsubscript{1}, d\textsubscript{2}\rangle\rightarrow d\textsubscript{3}$.  

We adopt this system as our ``retrieval'' knowledge source and re-use the retrieved contexts that are provided, both for \textit{RATD} datasets and for each evaluation dataset (section \ref{sec:retrieval}). \citet{Hartill2023-pf} also provide a Reasoning Model that is trained in a multitask manner on a large number of datasets including their \textit{RATD} datasets. We train two additional Reasoning models in the same manner as \citet{Hartill2023-pf} with, and without, the \textit{RATD} datasets (section \ref{sec:reasoning_models}). By reusing all of the above components we are able to quantify the effect of adding the second knowledge source under both the \textit{RR} and \textit{RATD} methods versus the baselines established by \citet{Hartill2023-pf} (section \ref{sec:experiments}).

\subsection{Rationale Generation}

We utilize two LLMs, BLOOM \cite{BigScience_Workshop2022-dh} and StableVicuna \citep{Stability-AI2023-mp}, a much smaller model that has been further tuned from the Vicuna v0 13B model \citep{Chiang2023-xe} which in turn was adapted from the LLama \cite{Touvron2023-nj} foundation model. We chose these two models because they are representative of differing approaches to developing LLMs and they may offer divergent characteristics in rationale generation. At 176 billion parameters, BLOOM is the largest language model we had access to at the time that we could run under INT8. It is trained on 410 billion tokens and the version we used did not undergo further training on instructional data or human feedback. Llama by contrast is trained on one trillion tokens. From the Llama checkpoint, Vicuna underwent further training on user-provided ChatGPT conversations. Finally StableVicuna was developed from Vicuna by further training in both supervised and reinforcement learning from human feedback (RLHF) \cite{Ouyang2022-ti} settings on a mixture of the human-generated OpenAssistant Conversations Dataset \cite{Kopf2023-bx} and human-LLM conversations from the GPT4All \cite{Anand2023-ma} and Alpaca \cite{Taori2023-au} projects. We used StableVicuna under both INT8 and FP16 versions, the former offering a smaller GPU memory footprint at around 18GB while the latter uses almost twice as much memory but we find inference much faster, thus offering a clear trade-off in a resource-constrained setting.

To generate rationales from each model, we used greedy decoding on chain-of-thought (COT) prompts \citep{Wei2022-lz} to generate the rationale followed by the phrase ``So the answer is'' and the answer (examples are in appendix \ref{sec:app_llmrationaleprompts}). This enabled us to evaluate the LLM answers directly from the same prompts and with the same rationale that our Reasoning Model would use, allowing a comparison under a similar set of assumptions. Occasionally a model would fail to generate the separate answer. In this case, to be favorable to the direct LLM method, the full rationale was used as the answer in calculating metrics. Generated rationale length is a maximum of 128 tokens.

To maintain the integrity of our unseen settings we ensured that no examples used in prompts were from any of our evaluation datasets. The prompts used were identical between our LLMs excepting that examples for StableVicuna prompts are denoted as:

\texttt{\#\#\# Human:\ [question]\ \#\#\# Assistant:\ [rationale].\ So the answer is [answer].}

BLOOM prompts are denoted as:

\texttt{Q:\ [question]\ A:\ [rationale].\ So the answer is [answer].} \\

Our qualitative examination of rationales generated by BLOOM and StableVicuna suggests a diversity in quality from both models but that they tend to produce better rationales on the same datasets (e.g. ARC-DA) and worse on the same (e.g. Musique). We observed that BLOOM was generally more prone to generating falsehoods. Examples from both models may be found in appendix \ref{app:eval_rationale_examples}. We note that robust examination of rationale quality is presently challenging to perform and believe research into automated methods in this area represents a promising future direction.

\subsection{Retrieval}
\label{sec:retrieval}

As well as the $n$-hop retrieval model discussed above, the Iterator also comprises a two-stage reranking system. The first stage is an $n$-hop Paragraph Reranker that scores retrieved paragraphs and sentences within paragraphs for relevance to the query at the current hop e.g. input $\langle q, d\textsubscript{0}, d\textsubscript{1}\rangle \text{ to score } d\textsubscript{1}$ on hop 2. Top-scoring sentences are passed to a second stage Evidence Set Scoring model that re-scores each sentence in the context of the accumulated set of top-scored sentences to the current hop (Evidence Set) as well as scoring the overall relevance of the Evidence Set.

For our ``retrieval'' knowledge source, as noted we simply reuse contexts generated by the Iterator, both for each evaluation sample and also for the creation of \textit{RATD} datasets for the training regimes. Iterator-generated contexts are formatted as a list of paragraph fragments that are recovered from the top-scored sentences, each prepended by the title of the corresponding document and containing the top-scoring sentences along with preceding and successor sentences where these exist. The top-scored sentences are identified by taking the Evidence Set from the top-scored hop. Contexts contain as many fragments as will fit into a 512-token sequence length. They are semi-structured as follows:

\texttt{[Doc 1 title]:\ [One to three sentences from document 1 paragraph].\ [Doc 2 title]:\ ...} \\

The corpus utilised by the Iterator is obtained from the August 1 2020 English Wikipedia dump and consists of approximately 35 million paragraphs.

\subsection{Rationale Ranker}

Our \textit{RR} model takes a question and context pair as input $\langle q, c\rangle$ and produces a score $s$. It is trained with a binary cross-entropy objective where samples are labelled 1.0 if $c$ is truthful and fully evidential in answering $q$ or 0.0 otherwise. The model is trained on a mixture of existing datasets for which we acquire or construct positive $c$ (i.e. a set of relevant and truthful gold sentences that are sufficient to answer $q$), and negative $c$ (which omit some or all gold sentences and may be irrelevant, false or both with respect to $q$ answerability). We used shared normalization \citep{Clark2018-oq} such that each $q$ is sampled in the same batch paired with a positive and separately a negative $c$. We found that without shared normalization, model training would collapse and it would predict every $c$ as negative. This may have occurred because without seeing positive and negative $c$ for the same $q$ in the same batch the pattern to be learned is insufficiently signalled.   

\begin{table}[h]
\centering
\caption{\textit{RR} model training dataset composition. The construction methods denoted ``... facts'' involve creating rationales from gold sentences or structured triples sourced from the cited study. Iterator-like contexts and Rationale-like are constructed from the training datasets' gold (and associated negative) paragraphs. LLM-sampled and LLM-greedy contexts are negative rationales generated by BLOOM using nucleus sampling and greedy decoding respectively. \textsuperscript{\textit{a}}\citet{Onoe2021-il}; \textsuperscript{\textit{b}}\citet{Yang2018-xq}; \textsuperscript{\textit{c}}\citet{Thorne2018-ws}; \textsuperscript{\textit{d}}\citet{Khot2020-sv}; \textsuperscript{\textit{e}}\citet{Clark2016-xg, Clark2018-gy}; \textsuperscript{\textit{f}}\citet{Jiang2020-on}; \textsuperscript{\textit{g}}\citet{Inoue2020-by}; \textsuperscript{\textit{h}}\citet{DeYoung2020-cx}; \textsuperscript{\textit{i}}\citet{Jhamtani2020-dh}; \textsuperscript{\textit{j}}\citet{Xie2020-xb}}
\label{tab:rr_train_datasets}
\resizebox{\textwidth}{!}{%
\begin{tabular}{l|rl|rl}
\toprule
 & & \textbf{Positive Contexts} & & \textbf{Negative Contexts} \\
\textbf{Training Mixture} & \textbf{Count} & \textbf{Construction Methods} & \textbf{Count} & \textbf{Construction Methods} \\ \midrule
Creak\textsuperscript{\textit{a}} (Commonsense) & 10173 & Creak facts\textsuperscript{\textit{a}} & 81408 & LLM-sampled \\
HotpotQA\textsuperscript{\textit{b}} (Multi-hop factual) & 34304 & R4C facts\textsuperscript{\textit{g}}, Iterator-like, Rationale-like & 41839 & LLM-sampled, LLM-greedy, Iterator-like, Rationale-like \\
FEVER\textsuperscript{\textit{c}} (Single-hop factual) & 60986 & Eraser facts\textsuperscript{\textit{h}}, Iterator-like, Rationale-like & 121427 & LLM-sampled, Iterator-like, Rationale-like \\
QASC\textsuperscript{\textit{d}} (Multi-choice science) & 47830 & QASC facts\textsuperscript{\textit{d}}, eQASC facts\textsuperscript{\textit{i}} & 193214 & LLM-sampled, LLM-greedy \\
ARC\textsuperscript{\textit{e}} (Multi-choice science) & 6469 & WorldTree facts\textsuperscript{\textit{j}} & 24492 & LLM-sampled, LLM-greedy \\
Hover\textsuperscript{\textit{f}} (Multi-hop factual) & 28171 & Iterator-like, Rationale-like & 28171 & Iterator-like, Rationale-like \\ \midrule
\textbf{Total} & \textbf{187933} & \textbf{} & \textbf{490551} & \textbf{} \\ \bottomrule
\end{tabular}%
}
\end{table}

Since the model must score both rationale-style $c$ and Iterator-generated $c$ on the same scale, we develop training samples that are similar to both types. Obtaining positive $c$ for training questions is generally straightforward. These are constructed from gold sentences and paragraphs associated with each dataset. Negative $c$ that cover both irrelevance and falsehood are harder to obtain. We construct negative $c$ by two methods; (1) generating them from BLOOM using specially constructed few-shot prompts containing examples of negative rationales (e.g. appendix \ref{app:neg_rationale_prompt}), and (2) creating them synthetically by substituting gold sentences with negative ones using datasets such as HotpotQA that come with sentence level annotations. The synthetic method can only produce irrelevant negatives whereas the LLM-generating method produces both irrelevant and false rationales. For LLM generation we use both greedy decoding and nucleus sampling \citep{Holtzman2019-nw} to create negatives. We find that greedy decoding produces positive-appearing but negative samples but (obtusely) the LLM has a tendency to produce accidentally positive rationales which we must filter out\footnote{We eliminate rationales where the stemmed text contains the stemmed answer string, excepting samples with yes/no labels. We use the snowball stemmer from NLTK \citep{Bird2009-cm}.}. Nucleus sampling by contrast (temperature=0.95 and p=0.96) produces a diversity of false and irrelevant samples that are less likely to be accidental positives. However here falsehoods tend to have an exaggerated quality which could make them less adversarial for the model, so we create samples via both decoding methods (examples in appendix \ref{app:neg_rationale_examples}). Dataset construction is summarised in Table \ref{tab:rr_train_datasets}.

We employ diverse combination methods involving the trained \textit{RR} model scores to create contexts for our evaluation datasets that combine rationales and Iterator-generated contexts, as described in section \ref{sec:context_combination_strategies}. 

\subsubsection{Rationale Ranker Evaluation}

Our \textit{RR} development set consists of 89,470 samples taken from the respective development splits of our training datasets. Contexts are created using the same methods as illustrated in Table \ref{tab:rr_train_datasets} for corresponding training splits. We sample a single positive or negative context for each development question such that there are equal positive and negative contexts. As shown in Table \ref{tab:rr_indomain_acc}, accuracy is high in this in-domain setting.

\begin{table}[h]
\centering
\caption{\textit{RR} model Accuracy on the in-domain development set (score threshold $t=0.5$). Total is micro-accuracy. High accuracy is attainable in detecting both positive and negative contexts.}
\label{tab:rr_indomain_acc}
\begin{tabular}{@{}rrr@{}}
\toprule
\textbf{Positive Context} & \textbf{Negative Context} & \textbf{Total} \\ \midrule
91.5 & 93.0 & 92.3 \\ \bottomrule
\end{tabular}
\end{table}

\begin{table}[h]
\centering
\caption{Accuracy in detecting falsehoods on TruthfulQA MC1. The \textit{RR} model is better at detecting falsehoods than the Iterator Paragraph Reranker which was trained to detect relevance but not falsehood. It's performance is competitive or better than much larger models that have not been trained using RLHF \textsuperscript{\textit{a}}\citet{OpenAI2023-cv}; \textsuperscript{\textit{b}}from \citet{Lin2022-bq} Github repository; \textsuperscript{\textit{c}}model from \citet{Hartill2023-pf} with results calculated by us.}
\label{tab:rr_truthfulqa_sqa}
\begin{tabular}{@{}lrr@{}}
\toprule
\textbf{Model} & \textbf{\begin{tabular}[c]{@{}r@{}}TruthfulQA\\ MC1\end{tabular}} &  \\ \midrule
GPT-4 RLHF\textsuperscript{\textit{a}} & 60.0 &  \\
GPT-3.5 RLHF\textsuperscript{\textit{a}} & 47.0 &  \\ \hdashline
GPT-4 No RLHF\textsuperscript{\textit{a}} & 30.0 &  \\
GPT-3 175B\textsuperscript{\textit{b}} & 21.0 &  \\
GPT-J 6B\textsuperscript{\textit{b}} & 20.0 &  \\
UnifiedQA 3B\textsuperscript{\textit{b}} & 19.0 &  \\ \midrule
Iterator Paragraph Reranker 335M\textsuperscript{\textit{c}} & 18.2 &  \\
Rationale Ranker 335M (Ours) & 30.0 &  \\ \bottomrule
\end{tabular}
\end{table}

Turning to an unseen setting, we initially evaluate context relevance scoring with a five-way multi-choice relevance detection dataset that we create from the gold rationales supplied with StrategyQA (SQA), where the four incorrect options are simply randomly assigned rationales from other SQA questions (we use SQA since this is not part of \textit{RR} model training). Here our model achieves 91.4\% accuracy. A more interesting question is the extent to which our relatively small \textit{RR} model is capable of detecting falsehoods in an unseen setting. To evaluate this question we consider TruthfulQA \citep{Lin2022-bq}, an adversarial evaluation-only dataset of 817 questions that models and/or humans tend to answer falsely. In Table \ref{tab:rr_truthfulqa_sqa} we compare falsehood detection performance of the \textit{RR} model with various larger models and in particular with the Iterator Paragraph Reranker. We treat the Paragraph Reranker as representative of models specifically trained to score context relevance but that have not necessarily been trained to consider truthfulness. We utilise the TruthfulQA MC1 split which is formatted as 4-5 way multi-choice with one truthful option. Each option is scored independently of other options and the highest-scoring selected as the prediction. In the case of LLMs the score is calculated as the log-probability of the completion following the question. For the Paragraph Reranker and our \textit{RR} model we use the score that each model has been trained to compute. It can be seen that the \textit{RR} model is indeed much better at detecting falsehoods than the Paragraph Reranker and it's performance is competitive or better than much larger models that have not been trained using RLHF. We imagine the superior performance of LLMs trained with RLHF on falsehood detection is due to their associated large reward models, like our \textit{RR} model, being trained in part to rate samples making false assertions as undesirable.

\subsection{Reasoning Models}
\label{sec:reasoning_models}

We consider three Reasoning Models in our experiments. Reasoning Models take a question and context pair as input ⟨q, c⟩ and generate an answer $a$. The first, which we use as a baseline, is the unmodified \textit{``Base+RATD''} model from \citet{Hartill2023-pf} which we denote here as the \textit{RATD} model for brevity. This is a multitask-trained model which is further trained from the original BART \citep{Lewis2020-gt} pretrained checkpoint on a large number of datasets\footnote{We refer the reader to \citet{Hartill2023-pf} for a more exhaustive description of the training regime and dataset construction.}. For descriptive purposes, we divide these training datasets into two sets. The first are the \textit{RATD} datasets described in section \ref{sec:retrieval}, whose purpose is to confer an ability to reason over long, noisy, and partially evidential contexts. We denote the remaining large number of training datasets as the \textit{Common} set; these broadly cover tasks designed to instill simple numerical literacy, and diverse question-answering ability. Hence we say that the \textit{RATD} model is trained on $\textit{Common} \cup \textit{RATD}$ datasets. 

We create an additional set of training samples denoted \textit{GR} (for ``gold rationales''). These are intended to impart further ability to reason over rationale-form contexts. \textit{GR} consists of samples for Creak, QASC, ARC, HotpotQA, and FEVER where the contexts are gold rationales constructed similarly and from the same sources as those described for the \textit{RR} model training dataset in Table \ref{tab:rr_train_datasets}. 

We then develop our two main Reasoning Models, both multitask-trained using the same approach and hyperparameters as the original \textit{RATD} model: The \textit{GR} model is trained on $\textit{Common}\ \cup\ \textit{GR}$, and the \textit{GR+RATD} model is trained on $\textit{Common}\ \cup\ \textit{GR}\ \cup\ \textit{RATD}$.

\section{Experiments}
\label{sec:experiments}

%

\subsection{Models}
The Rationale Ranker is built upon ELECTRA-large \citep{Clark2020-vh}. Reasoning Models are based on BART \citep{Lewis2020-gt}. All models use the the Huggingface \citep{Wolf2020-ro} implementations. The Reasoning Models differ only in their respective training data; hyperparameters are otherwise identical.

\subsection{Unseen Evaluation Datasets}
\label{sec:unseen_eval_datasets_desc}

All evaluation dataset results are reported against the same splits  used by \citet{Hartill2023-pf}. As with that paper we use the numeracy-focused F1 calculation introduced in \citet{Dua2019-td} for ARC-DA, IIRC and Musique.

\textbf{StrategyQA} \citep{Geva2021-sl} (SQA) contains commonsense samples involving diverse multi-hop reasoning strategies with yes/no answers (average $n = 2.33$). The full training set is used for evaluation as with BIG-bench \citep{Srivastava2022-kp}.

\textbf{CommonsenseQA} \citep{Talmor2019-rm} (CSQA) is a multi-choice dataset of commonsense questions derived from Conceptnet \citep{Speer2017-ll}. The task is to choose the best option from five options of which more than one may sometimes be plausible.

\textbf{IIRC} \citep{Ferguson2020-hv} contains factual questions and an initial explanatory paragraph for each which must be augmented with additional retrieved information to be fully evidential ($1 \leq n \leq 4+$). Answers may be numbers, binary, text spans or labeled unanswerable.

\textbf{ARC-DA} \citep{Bhakthavatsalam2021-fq} is a subset of ARC \citep{Clark2018-gy} (science questions) where questions have been re-worded to make sense in an open domain context. The original multichoice versions of ARC are part of our training regime for both Reasoning and \textit{RR} models, so samples are ``partially unseen'' in the sense that the question format is different.

\textbf{Musique} \citep{Trivedi2022-mv} is a $n$-hop factual dataset ($n \leq 4$) constructed by combining single-hop questions from existing datasets. The training split of Musique is used in all of our Reasoning Models, and in the Iterator training. However as with \citet{Hartill2023-pf}, we use the original development split as ``partially seen'' since development samples were constructed such that no single hop question, answer, or associated paragraph is common to the corresponding element of any training sample. Hence the form of questions is ``seen'' but the exact questions are not.

\subsection{Context Combination Methods and Experimental Nomenclature}
\label{sec:context_combination_strategies}

\begin{figure*}[h]
\begin{center}
\includegraphics[width=\textwidth]{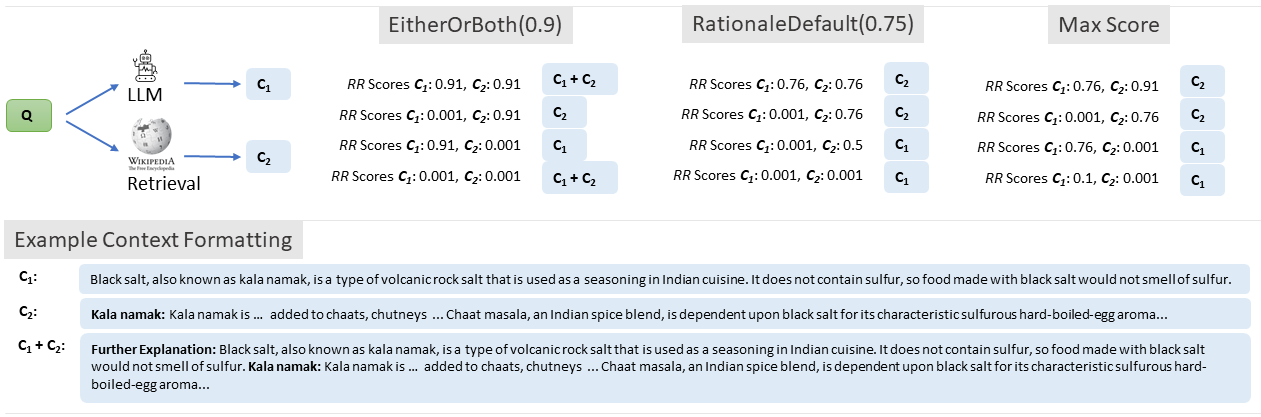}
\caption{Examples of combining contexts. For a question \textbf{Q}, we acquire two contexts, \textbf{C\textsubscript{1}} and \textbf{C\textsubscript{2}}. The resulting combined context for our combination methods with example thresholds and \textit{RR} model scores is then shown in blue boxes where ``+'' denotes the concatenation of \textbf{C\textsubscript{1}} and \textbf{C\textsubscript{2}}. The Naïve Concatenation is always \textbf{C\textsubscript{1}} + \textbf{C\textsubscript{2}}. Formatted examples of resulting contexts are shown at the bottom of the figure with titles shown in bold for readability. The phrase ``Further Explanation'' is added to the rationale in a concatenated context to mimic a document title. }
\label{fig_combo_examples}
\end{center}
\end{figure*} 

For each unseen evaluation question, given a LLM-generated rationale, and an Iterator-generated context as possible combined context components, and \textit{RR} model scores for each, we evaluate methods of combining components. We implement four combination methods and create versions of our unseen evaluation datasets with combined contexts for each as follows: 

\textbf{Naïve Concatenation}: The simple concatenation of a rationale and corresponding Iterator-generated context with the above form. \textit{RR} model scores are ignored.

\textbf{Max Score}: Choosing the single component that the \textit{RR} model scores highest.

\textbf{RationaleDefault}: Defaulting to taking the rationale component unless the Iterator component scores over a threshold $t$ in which case it is exclusively selected.

\textbf{EitherOrBoth}: Selecting either or both components that score over a threshold $t$. If neither component is selected, we default to the Naïve Concatenation context since smaller Language Models have been shown to be ineffective for answering unmemorized question-only (open domain) questions \citep{Lewis2021-ia}.

For the latter two combination methods we create contexts using each of eight \textit{RR} score thresholds ranging from $t=0.0005$ to $t=0.9$. We denote the particular version using the threshold e.g. EitherOrBoth(0.9) means samples are augmented using the EitherOrBoth method with $t=0.9$. Obviously innumerably other combination methods are possible but we find that this set is sufficient for our research purposes while remaining manageable. Figure \ref{fig_combo_examples} illustrates examples of contexts derived from each combination method using hypothetical \textit{RR} scores. Combined contexts are truncated (from the Iterator component) to the maximum sequence length of the model (512 tokens) at inference time.

Each of our three Reasoning Models might be expected to perform better with particular context types. For example the \textit{GR} model might do better where the context tends to be rationale-like whereas the \textit{RATD} model may do better where the context is of Iterator-generated form. This influences which combination method is likely to perform better on each Reasoning Model.

Similarly, different combination methods are likely to work better for differing question types (commonsense, multi-hop factual etc). For example knowing that LLM-generated rationales tend to be more effective than Iterator-generated contexts for answering commonsense questions, we can deduce that RationaleDefault(0.9) is likely to be a good strategy for developing contexts for CommonsenseQA because using this strategy results in Rationale-only contexts except where the Iterator context is scored very highly. However, we are interested in the situation where our model is presented with an arbitrary question of unknown type. Hence we are more interested in finding combination methods that will \textit{generally} work well under this assumption, even where the method may not be the best for any particular type. We identify combination methods satisfying this criteria as those with the highest \textit{unweighted macro-average score over our unseen evaluation datasets} (henceforth ``Mean'' or ``Mean score'') on each Reasoning Model, taking inspiration for averaging over heterogeneous metrics from e.g. \citet{Wang2019-ll,Wang2019-gf}. For the methods that utilize \textit{RR} model scores we select the highest performing on this measure and refer to it as ``Generally best RR combo'' below. We also report the ``Best RR combo per dataset'' where we select the highest scoring combination method for each evaluation dataset. We note that since we cannot use this approach on an arbitrary question of unknown type we don't consider it a usable method in a truly unseen setting, although future work could remedy this (e.g. through utilising an additional model trained to predict the best combination method for a question). 

We refer below to contexts created for each evaluation dataset that consist entirely of Iterator-generated contexts as ``Iterator only'', those contexts entirely composed of LLM-generated rationales as ``Rationale only'', and those that apply any of the combining methods as ``Rationale + Iterator'' (noting that individual samples in the latter may only contain one of the possible context components). For brevity, where referring to the use of a particular context type on a particular model we use shorthand such as ``\textit{GR+RATD}: Iterator only'' or ``\textit{GR+RATD}: Iterator + Rationale (Naïve Concatenation)''.

To test statistical significance over the large number of model:context combinations created we use methods for accomplishing this described in \citet{Demsar2006-ze} as implemented in the AutoRank library \citep{Herbold2020-vn}. Specifically all tests use significance level $\alpha=0.05$ and we use the non-parametric Friedman test as omnibus test, followed by the Nemenyi test to infer which differences are significant. Significance test results are summarised in Appendix \ref{sec:app_stat_tests}.

\subsection{Experimental Results}

\subsubsection{Summary}
\begin{table}[h]
\centering
\caption{Mean score over unseen evaluation datasets. The ``Iterator only'' results are duplicated across Rationale Generators to facilitate comparison. Bold indicates highest score per context type (i.e. per row). StableVicuna-generated rationales generally outperform BLOOM rationales.}
\label{tab:main_1_model_contexttype_overview}
\resizebox{\textwidth}{!}{%
\begin{tabular}{l|rrr|rrr}
\toprule
\textbf{Rationale Generator $\rightarrow$ } & \multicolumn{3}{|c}{\textbf{StableVicuna (INT8)}} & \multicolumn{3}{|c}{\textbf{BLOOM (INT8)}} \\
\textbf{Context $\downarrow$ / \textit{Model} $\rightarrow$} & \textbf{\textit{GR}} & \textbf{\textit{RATD}} & \textbf{\textit{GR+RATD}} & \textbf{\textit{GR}} & \textbf{\textit{RATD}} & \textbf{\textit{GR+RATD}} \\ \midrule
Iterator only & 38.1 & 40.4 & \textbf{41.0} & 38.1 & 40.4 & \textbf{41.0} \\
Rationale only & 44.5 & 44.2 & \textbf{45.3} & 39.5 & 42.0 & 40.3 \\
Rationale + Iterator (Naïve concatenation) & 42.7 & 46.3 & \textbf{47.2} & 43.2 & 43.8 & 43.7 \\
Rationale + Iterator (Generally best RR combo) & 45.5 & 46.3 & \textbf{47.2} & 42.9 & 44.2 & 44.4 \\ \midrule
Rationale + Iterator (Best RR combo per dataset) & 47.6 & 47.5 & \textbf{48.1} & 45.1 & 45.6 & 45.4 \\ \bottomrule
\end{tabular}%
}
\end{table}

As Table \ref{tab:main_1_model_contexttype_overview} indicates, rationales generated by BLOOM almost always produce weaker results than those from StableVicuna. For example, in considering BLOOM-generated ``Rationale only'' contexts, the \textit{GR} model might have been expected to outperform the \textit{RATD} model (given the additional samples with gold rationale contexts added to \textit{GR} training). However the \textit{GR} model actually underperforms (39.5 vs 42.0). Conversely, where considering StableVicuna-generated ``Rationale only'' contexts, the \textit{GR} model slightly outperforms the \textit{RATD} model as expected.

\subsubsection{\textit{GR+RATD} Model Versus Baseline And LLM Direct Prompts}
\label{gr_ratd_vs_llm}

It can be seen in Table \ref{tab:main_1_model_contexttype_overview} that where using the stronger StableVicuna-generated rationales, the \textit{GR+RATD} model results dominate both \textit{RATD} and \textit{GR} models, so we consider this as our best model. Table \ref{tab:main_2_per_dataset} compares \textit{GR+RATD} to our main baseline (i.e. ``\textit{RATD}: Iterator only'' from \citet{Hartill2023-pf}). Both our ``Naïve concatenation'' and ``Generally best RR combo'' combination methods significantly outperform this baseline on the Mean score and on most individual datasets, except for Musique.

\begin{table}[h]
\centering
\caption{Evaluation per dataset. The ``Rationale+Iterator'' combined contexts significantly outperform the ``\textit{RATD}: Iterator only'' baseline and both single-component contexts. The ``Rationale only'' row using StableVicuna-generated rationales significantly outperforms the StableVicuna COT direct prompt. Bold indicates best in column excluding Best Prior and Best RR combo per dataset. Best prior are either not  unseen or involve much larger models as follows: \textsuperscript{\textit{a}}\citet{Anil2023-ws}: Palm 2 using self consistency. \textsuperscript{\textit{b}}\citet{Xu2021-ol}: Finetuned, retrieval from Conceptnet. \textsuperscript{\textit{c}}\citet{Bhakthavatsalam2021-fq}: Training includes ARC-DA. \textsuperscript{\textit{d}}\citet{Hartill2023-pf}: Finetuned. \textsuperscript{\textit{e}}\citet{Trivedi2022-mv}: Specialised retrieval from gold and distractor paragraphs.}
\label{tab:main_2_per_dataset}
\resizebox{\textwidth}{!}{%
\begin{tabular}{lrrrrrrr}
\toprule
\textbf{\textit{Model}: Context} & \textbf{SQA} & \textbf{CSQA} & \textbf{ARC-DA} & \textbf{IIRC} & \textbf{Musique} & \textbf{Mean} \\
 & \textbf{(Acc.)} & \textbf{(Acc.)} & \textbf{(F1)} & \textbf{(F1)} & \textbf{(F1)} & \\ \midrule
Random &  50.0 & 20.0 &  &  &  &  \\
Best Prior &  90.4\textsuperscript{\textit{a}} & 91.2\textsuperscript{\textit{b}} & 61.4\textsuperscript{\textit{c}} & 53.6\textsuperscript{\textit{d}} & 49.8\textsuperscript{\textit{e}} & 69.3 \\ \midrule
\textit{RATD}: Iterator only & 58.9 & 63.6 & 31.6 & 25.5 & \textbf{22.2} & 40.4 \\ \hdashline
\textit{BLOOM INT8}: Few Shot Standard Prompt  & 58.1 & 47.5 & \textbf{58.7} & 17.3 & 9.4 & 38.2 \\
\textit{StableVicuna INT8}: Few Shot Standard Prompt & 56.2 & 70.8 & 56.8 & 19.8 & 9.3 & 42.6 \\ \hdashline
\textit{BLOOM INT8}: Few Shot COT Prompt & 57.1 & 54.9 & 50.5 & 17.4 & 11.1 & 38.2 \\
\textit{StableVicuna INT8}: Few Shot COT Prompt  & 61.7 & 67.7 & 45.8 & 20.8 & 12.6 & 41.7 \\ \hdashline
\textit{GR+RATD}: Iterator only & 57.3 & 65.0 & 35.6 & 25.6 & 21.5 & 41.0 \\
\textit{GR+RATD}: Rationale only & \textbf{64.2} & \textbf{73.1} & 50.2 & 25.1 & 13.8 & 45.3 \\
\textit{GR+RATD}: Rationale + Iterator (Naïve concatenation) & 61.7 & 72.6 & 53.0 & 27.0 & 21.7 & \textbf{47.2} \\
\textit{GR+RATD}: Rationale + Iterator (Generally best RR combo) & 61.7 & 72.7 & 52.1 & \textbf{27.3} & 22.0 & \textbf{47.2} \\ \midrule
\textit{GR+RATD}: Rationale + Iterator (Best RR combo per dataset) & 64.5 & 73.3 & 53.0 & 27.4 & 22.4 & 48.1 \\ \bottomrule
\end{tabular}%
}
\end{table}

We next consider the efficacy of directly prompting both LLMs to produce the answer using few-shot COT exemplars, and separately with standard few-shot prompts that use the same exemplars without the rationale portions. Here, the most like-for-like comparison is from the StableVicuna COT prompt to ``\textit{GR+RATD}: Rationale only'', since the rationales used are the same ones produced by the direct StableVicuna COT prompts. For the StableVicuna COT prompt (and both BLOOM prompts), ``\textit{GR+RATD}: Rationale only'' significantly outperforms the LLM direct prompts on the overall Mean score, and generally on individual datasets (except for ARC-DA). The 42.6 to 45.3 Mean improvement is not significant for the StableVicuna Standard prompt. 

In comparing performance of our combined contexts (``Naïve concatenation'' and ``Generally best RR combo'') to the single-component contexts (``Iterator only'' and ``Rationale only''), both combined contexts achieve a higher Mean score than either single component context does (improvement from ``Iterator Only'' is significant in both cases, that from ``Rationale Only'' to ``Naïve concatenation'' is significant, the other is on the significance threshold (appendix \ref{tab:app:significance_tests})). Notably, three of the five datasets (ARC-DA, IIRC and Musique) have higher scores on either combined context than on any single component context as well. 

Considering the ``Iterator only'' against the ``Rationale only'' rows in Table \ref{tab:main_2_per_dataset} illuminates the relative strengths of our two knowledge sources. Multi-hop factual questions as exemplifed in Musique benefit far more from retrieved paragraphs than LLM-generated rationales (21.5 F1 vs 13.8 F1) whereas commonsense datasets such as SQA (64.2 acc vs 57.2 acc) and CSQA (73.1 acc vs 65.0 acc) unsurprisingly benefit more from LLM-generated rationales as context. IIRC, another factual dataset might have been expected to benefit more from retrieved paragraphs but performance is similar between rationale-only contexts and retrieved paragraphs. We suggest this is because the input for each IIRC sample is comprised of the question and the initial gold paragraph, and many samples then only require a single extra piece of information in order to have sufficient evidence. LLMs may be better at performing (the equivalent of) this single hop than they are at identifying the multiple additional pieces of information necessary in the Musique case.

\subsubsection{\textit{RR} Model Scoring And \textit{RATD} Training Efficacy}
\label{rr_ratd_improvement quantification}

We next evaluate the effectivness of our methods through an ablational approach. The \textit{GR} model can be regarded as an ablation of \textit{RATD} training from the \textit{GR+RATD} model (-RATD). The Naïve concatenation context type can be seen as an ablation of \textit{RR} Model scoring from the Generally best RR combo (-RR). Hence our ``\textit{GR}: Rationale + Iterator (Naïve concatenation)'' model can be seen as an ablation of both (-RR -RATD) while being (insignificantly) better than the main ``RATD: Iterator only'' baseline (40.4 vs 42.7). Table \ref{tab:ablations_rr_ratd} illustrates the relative efficacy of our two methods, both individually and together. What is revealed is that the \textit{RR} model-scoring approach significantly improves Mean results in the absence of \textit{RATD} training (45.5 vs 42.7), while the \textit{RATD} training significantly improves results in the absence of \textit{RR} scoring (47.2 vs 42.7). The difference between the two methods (45.5 vs 47.2) is \textit{not} significant. 

\begin{table}[h]
\centering
\caption{\textit{RATD} and \textit{RR} effectiveness. The bottom row can be regarded as an ablation of both \textit{RR} and \textit{RATD} (-RR -RATD). All three topmost methods (marked with an asterisk) are significantly different from the bottow row (-RR -RATD) however differences between the three topmost methods are \textit{not} significant. This shows that the \textit{RR} and \text{RATD} methods are individually both effective but combining the methods does not improve results further.}
\label{tab:ablations_rr_ratd}
\begin{tabular}{@{}llr@{}}
\toprule
\textbf{\textit{Model}: Context} & & \textbf{Mean} \\ \midrule
\textit{GR+RATD}: Rationale + Iterator (Generally best RR combo) & +RR +RATD\textsuperscript{*} & 47.2 \\
\textit{GR+RATD}: Rationale + Iterator (Naïve concatenation) & -RR +RATD\textsuperscript{*} & 47.2 \\
\textit{GR}: Rationale + Iterator (Generally best RR combo) & +RR -RATD\textsuperscript{*} & 45.5 \\
\textit{GR}: Rationale + Iterator (Naïve concatenation) & -RR -RATD & 42.7 \\ \bottomrule
\end{tabular}
\end{table}

Using the two methods in combination does not improve results further. The ``Generally best RR combo'' for the \textit{GR+RATD} model uses the EitherOrBoth(0.9) combination method. This can be interpreted as only selecting a context component if the \textit{RR} model scores it very highly, and since both components frequently fail to meet the threshold the default of using the Naïve concatenation then applies. This has the effect of the context being the Naïve concatenation for 80.9\% of evaluation samples (Appendix \ref{sec:app_context_component_analysis}) which explains why combining the \textit{RATD} and \textit{RR} doesn't result in further improvement in this case.

\section{Conclusion}
\label{sec:bibtex}

We have implemented methods for combining explanatory context from two knowledge sources: LLM-generated rationales and retrieved paragraphs from Wikipedia. The first method involves training our smaller Reasoning Model on \textit{RATD} datasets such that it becomes proficient at reasoning over long, noisy contexts which contain information from both knowledge sources. The second method is to use Rationale Ranking model scores for each knowledge source as guidance in constructing contexts that may contain information from both, or either knowledge source. We have shown that both methods are individually effective in significantly improving unseen question-answering performance both versus the baselines established by \citet{Hartill2023-pf} and versus a baseline that ablates both \textit{RR} and \textit{RATD} methods (section \ref{rr_ratd_improvement quantification}).

We have shown that smaller Language Models can manifest comparable or stronger reasoning performance to LLMs when provided with the same knowledge to reason over that the LLM is capable of generating for itself. (section \ref{gr_ratd_vs_llm}).

After comparing results from question-answering using LLM-generated rationales as context with those using retrieved paragraphs we concluded that LLMs are weaker at surfacing the multiple pieces of information necessary to answer multi-hop factual questions, but stronger at generating rationales suitable for answering commonsense questions. Both knowledge sources are found to be effective for question types such as factual questions requiring a single additional piece of information  (section \ref{gr_ratd_vs_llm}).

In comparing performance of our combined contexts to the single-component contexts, the combined contexts achieve a higher Mean score over all unseen evaluation datasets than either single component context does. Individually, three of the five datasets (ARC-DA, IIRC and Musique) achieve higher scores when using combined contexts than on any single component context as well  (section \ref{gr_ratd_vs_llm}).

\subsubsection*{Broader Impact Statement}
Our Reasoning Models following the application of our methods are still capable of generating hallucinated, false and/or potentially offensive answers. Hence usage is most appropriate for research environments. 

Conversely, as \citet{Hartill2023-pf} note, latency, physical compute size, cost and energy efficiency are important considerations where smaller models offer material benefits. A diversity of applications exist in the broad domain of reasoning systems and due weight should be assigned to all factors in determining the most appropriate approach for a particular situation.


\bibliography{custom_v2}
\bibliographystyle{tmlr}

\appendix

\section{Hyperparameters}
\label{sec:app_hyperparams}

All models are trained on one Nvidia A100 80GB GPU using mixed precision and a linear learning rate decay schedule. Initial learning rates and other hyperparameters are shown in Table \ref{tab:hyperparams}. The optimiser used for the Rationale Ranker is Adam. All other models use AdamW. A maximum sequence length of 512 tokens was used for all trained models. The \textit{GR} and \textit{GR+RATD} Reasoning Model training starts from the \textit{Base} model checkpoint described in \citet{Hartill2023-pf}.

BLOOM loaded under INT8 with a batch size of one consumed approximately 200GB of GPU RAM. StableVicuna also under INT8 with a batch size of one consumed approximately 18GB.

\begin{table}[h]
\centering
\caption{Hyperparameters used for each model. Each training step is one batch input i.e the number of optimization steps is $Training Steps / GradientAccumulationSteps$. All final models are selected as the best model on the development set(s) over the specified number of training steps.}
\begin{tabular}{@{}lrrrr@{}}
\toprule
\textbf{Model} & \textbf{\begin{tabular}[c]{@{}r@{}}Initial\\ LR\end{tabular}} & \textbf{\begin{tabular}[c]{@{}r@{}}Batch\\ Size\end{tabular}} & \textbf{\begin{tabular}[c]{@{}r@{}}Grad.\\ Accum\end{tabular}} & \textbf{\begin{tabular}[c]{@{}r@{}}Train\\ Steps\end{tabular}} \\ \midrule
Rationale Ranker & 5e-5 & 24 & 8 & 188K \\
Reasoning Models & 2e-5 & 32 & 4 & 1M \\ \bottomrule
\end{tabular}%
\label{tab:hyperparams}
\end{table}

\section{Prompts For LLM Rationale Generation}
\label{sec:app_llmrationaleprompts}

Here we show the few-shot COT prompts used against the StableVicuna Model. Prompts used against BLOOM use the same examples and instructions and simply have ``Q:'' substituted for ``\#\#\# Human:'' (excepting the initial instruction which substitutes the empty string), and ``A:'' substituted for ``\#\#\# Assistant:''. The corresponding few-shot standard prompts are identical except that the examples have both the rationale and the answer key (``So the answer is'') removed. Rationales were constructed by truncating the generation at the first <newline> character and then truncating again at the answer key.

\subsection{Binary-labelled Datasets (SQA)}

\texttt{\#\#\# Human: Write a truthful, factual explanation to help answer the question, then answer the question. \\ \\
\#\#\# Human: A pupil can be either a student or part of an eye? \\
\#\#\# Assistant: A pupil is another word for student. The pupil is also a hole located in the center of the iris of the eye that allows light to strike the retina. Thus pupil can have either meaning. So the answer is yes. \\
\#\#\# Human: Greece is larger than mexico? \\
\#\#\# Assistant: Greece is approximately 131,957 sq km, while Mexico is approximately 1,964,375 sq km, making Mexico 1,389\% larger than Greece. So the answer is no. \\
\#\#\# Human: Glasses always fog up? \\
\#\#\# Assistant: Condensation occurs on eyeglass lenses when water vapor from your sweat, breath, and ambient humidity lands on a cold surface, cools, and then changes into tiny drops of liquid, forming a film that you see as fog. Your lenses will be relatively cool compared to your breath when the outside air is cold but not when the air is warm. So the answer is no. \\
\#\#\# Human: A fish is capable of thinking? \\
\#\#\# Assistant: Fish are more intelligent than they appear. In many areas, such as memory, their cognitive powers match or exceed those of ’higher’ vertebrates including non-human primates. Fish’s long-term memories help them keep track of complex social relationships. So the answer is yes. \\
\#\#\# Human: Is a 5 by 8 pool of bricks larger than an 8 by 5 pool of pillows? \\
\#\#\# Assistant: The 5 by 8 pool of bricks has an area of 40. The 8 by 5 pool of pillows also covers an area of 40. Thus, both pools are the same size. So the answer is no. \\
\#\#\# Human: When listed alphabetically, would the words baseball and demonstration appear in this order? \\
\#\#\# Assistant: Baseball begins with letter b and demonstration begins with letter d. Alphabetically, b comes before d, thus baseball comes before demonstration. So the answer is yes. \\
\#\#\# Human: Would Sally use a spaceship if she needed to get to London from Sydney quickly? \\
\#\#\# Assistant: Commerical spaceship transport doesnt exist yet. London is far from Sydney and planes are faster than trains or boats over long distances. Thus, Sally would use an existing transportation option like a plane. So the answer is no. \\
\#\#\# Human: A common effect of smoking lots of cigarettes in one’s lifetime is a higher than normal chance of getting lung cancer? \\
\#\#\# Assistant: Those who consistently averaged less than one cigarette per day over their lifetime had nine times the risk of dying from lung cancer than non-smokers. Among people who smoked between one and 10 cigarettes per day, the risk of dying from lung cancer was nearly 12 times higher than that of non-smokers. So the answer is yes. \\
\#\#\# Human: A rock is the same size as a pebble? \\
\#\#\# Assistant: A pebble is a clast of rock with a particle size of 4 to 64 millimetres based on the Udden-Wentworth scale of sedimentology. Pebbles are generally considered larger than granules (2 to 4 millimetres diameter) and smaller than cobbles (64 to 256 millimetres diameter). So the answer is no. \\
\#\#\# Human: \{QUESTION\} \\
\#\#\# Assistant:}

\subsection{Span or binary answers (ARC-DA, IIRC, Musique)}

\texttt{\#\#\# Human: Write a truthful, factual explanation to help answer the question, then answer the question. \\ \\
\#\#\# Human: Greece is larger than mexico? \\
\#\#\# Assistant: Greece is approximately 131,957 sq km, while Mexico is approximately 1,964,375 sq km, making Mexico 1,389\% larger than Greece. So the answer is no. \\
\#\#\# Human: Tools can be made of wood, iron, plastic amongst other things. Iron tools historically replaced what? \\
\#\#\# Assistant: Historically, iron tools were first used in the Iron Age. The Iron Age followed the Bronze Age. Thus, iron tools replacing bronze tools makes most sense. So the answer is bronze tools. \\
\#\#\# Human: Which magazine was started first Arthur's Magazine or First for Women? \\
\#\#\# Assistant: Arthur's Magazine was an American literary periodical first published 1844. First for Women is a woman's magazine that started in 1989. So the answer is Arthur's Magazine. \\
\#\#\# Human: Glasses always fog up? \\
\#\#\# Assistant: Condensation occurs on eyeglass lenses when water vapor from your sweat, breath, and ambient humidity lands on a cold surface, cools, and then changes into tiny drops of liquid, forming a film that you see as fog. Your lenses will be relatively cool compared to your breath when the outside air is cold but not when the air is warm. So the answer is no. \\
\#\#\# Human: Water flows downhill and tends to collect in low-lying areas such as valleys and holes. Lakes are usually large bodies of fresh water. Where is a lake likely to be found? \\
\#\#\# Assistant: The answer must be somewhere that a lot of water can collect. A valley is likely to be able to collect enough water. So the answer is a valley. \\
\#\#\# Human: Foxes are predators whose natural habitat is the forest. They have also been known to eat chickens from people's properties. The fox walked from the city into the forest, what was it looking for? \\
\#\#\# Assistant: The answer must be a reason for a fox to go into the forest. The forest is a fox’s natural habitat. So the answer is natural habitat. \\
\#\#\# Human: A fish is capable of thinking? \\
\#\#\# Assistant: Fish are more intelligent than they appear. In many areas, such as memory, their cognitive powers match or exceed those of ’higher’ vertebrates including non-human primates. Fish’s long-term memories help them keep track of complex social relationships. So the answer is yes. \\
\#\#\# Human: Musician and satirist Allie Goertz wrote a song about the "The Simpsons" character Milhouse, who Matt Groening named after who? \\
\#\#\# Assistant: Milhouse Mussolini van Houten is a fictional character featured in the tv series "The Simpsons" created by Matt Groening. He named the character after President Richard Nixon's middle name. So the answer is President Richard Nixon. \\
\#\#\# Human: What is the largest annual event held in the birthplace of the performer who sang Let Me Tell You About the Birds and the Bees? \\
\#\#\# Assistant: The Birds and the Bees was a 1964 single release by Jewel Akens. Jewel Akens was born in Houston, Texas, USA. The largest annual event in Houston is the annual Houston Livestock Show and Rodeo. So the answer is Houston Livestock Show and Rodeo. \\
\#\#\# Human: A common effect of smoking lots of cigarettes in one’s lifetime is a higher than normal chance of getting lung cancer? \\
\#\#\# Assistant: Those who consistently averaged less than one cigarette per day over their lifetime had nine times the risk of dying from lung cancer than non-smokers. Among people who smoked between one and 10 cigarettes per day, the risk of dying from lung cancer was nearly 12 times higher than that of non-smokers. So the answer is yes. \\
\#\#\# Human: Fred owns a number of things including a baseball bat, a laptop, a tablet and a briefcase. Fred works a long way from his home, so which electronic device would Fred prefer to walk to work with? \\
\#\#\# Assistant: Electronic devices include tablets and laptops. Tablets are lighter than laptops and lighter things are easier than heavier things to carry. Thus, a tablet is the easiest device for Fred to carry. So the answer is tablet. \\
\#\#\# Human: Chang Ucchin was born in korea during a time that ended with the conclusion of what? \\
\#\#\# Assistant: Chang Ucchin was born when Korea was still under Japanese colonial rule. Korea under Japanese rule began 1910 and ended at the conclusion of World War II in 1945. So the answer is World War II. \\
\#\#\# Human: A rock is the same size as a pebble? \\
\#\#\# Assistant: A pebble is a clast of rock with a particle size of 4 to 64 millimetres based on the Udden-Wentworth scale of sedimentology. Pebbles are generally considered larger than granules (2 to 4 millimetres diameter) and smaller than cobbles (64 to 256 millimetres diameter). So the answer is no. \\
\#\#\# Human: When did the birth state of Kevin Sessums become a right to work state? \\
\#\#\# Assistant: Kevin Sessums was born in 1956 in Forest, Mississippi. The Right to work law was adopted by Mississipi in 1954. So the answer is 1954. \\
\#\#\# Human: \{QUESTION\} \\
\#\#\# Assistant:}

\subsection{Multi-choice Datasets (CSQA)}

\texttt{\#\#\# Human: Write a truthful, factual explanation to help answer the question, then answer the question. \\ \\
\#\#\# Human: A common effect of smoking lots of cigarettes in one’s lifetime is what? Answer Choices: (A) poverty (B) low chance of lung cancer (C) good fitness (D) high chance of knee cancer (E) high chance of lung cancer \\
\#\#\# Assistant: Those who consistently averaged less than one cigarette per day over their lifetime had nine times the risk of dying from lung cancer than non-smokers. Among people who smoked between one and 10 cigarettes per day, the risk of dying from lung cancer was nearly 12 times higher than that of non-smokers. So the answer is high chance of lung cancer. \\
\#\#\# Human: Which magazine was started first? Answer Choices: (A) History channel (B) Youtube (C) Arthur's Magazine (D) Climbing (E) First for Women \\
\#\#\# Assistant: Arthur's Magazine was an American literary periodical first published 1844. First for Women is a woman's magazine that started in 1989. So the answer is Arthur's Magazine. \\
\#\#\# Human: How do you put on a sock? Answer Choices: (A) jump in (B) insert hand (C) put on head (D) insert foot (E) open \\
\#\#\# Assistant: Socks are worn on feet and they have an opening at one end. A foot must be inserted into the opening to put it on. Thus, of the choices only insert foot makes sense. So the answer is insert foot. \\
\#\#\# Human: After earning a lot in tips at his job, what would a waiter want to do next? Answer Choices: (A) do handstand (B) quit job (C) find another job (D) grow flowers (E) save the money \\
\#\#\# Assistant: Tips are money earned by waiting on tables. After earning money, people like to spend or save it. Thus, of the choices, the waiter would want to save the money. So the answer is save the money. \\
\#\#\# Human: Iron tools historically replaced what? Answer Choices: (A) bronze tools (B) wooden tools (C) uranium tools (D) plastic tools (E) eels \\
\#\#\# Assistant: Historically, iron tools were first used in the Iron Age. The Iron Age followed the Bronze Age. Thus, of the choices, iron tools replacing bronze tools makes most sense. So the answer is bronze tools. \\
\#\#\# Human: What mode of transport should Sally use if she needed to get to London from Sydney quickly? Answer Choices: (A) train (B) plane (C) spaceship (D) fast boat (E) slingshot \\
\#\#\# Assistant: Realistic modes of transport include trains, planes and boats. London is far from Sydney and planes are faster than trains or boats over long distances. Thus, of the realistic choices, planes are a faster way to travel. So the answer is plane. \\
\#\#\# Human: What can be used to warm up your home? Answer Choices: (A) refrigerator (B) flamethrower (C) heat pump (D) dog (E) blanket \\
\#\#\# Assistant: Warming a house should be done safely and efficiently. Heat pumps are safe and efficient. Thus, of the choices, heat pumps are the best way to heat a home. So the answer is heat pump. \\
\#\#\# Human: Fred works a long way from his home, so which electronic device would Fred prefer to walk to work with? Answer Choices: (A) laptop (B) briefcase (C) banana (D) tablet (E) car \\
\#\#\# Assistant: Electronic devices include tablets and laptops. Tablets are lighter than laptops and lighter things are easier than heavier things to carry. Thus, of the realistic choices, tablet is the easiest for Fred to carry. So the answer is tablet. \\
\#\#\# Human: What activity is a fish is capable of? Answer Choices: (A) thinking (B) jogging (C) using tools (D) flight (E) dentistry \\
\#\#\# Assistant: Fish are more intelligent than they appear. In many areas, such as memory, their cognitive powers match or exceed those of ’higher’ vertebrates including non-human primates. Fish’s long-term memories help them keep track of complex social relationships. So the answer is thinking. \\
\#\#\# Human: Chang Ucchin was born in korea during a time that ended with the conclusion of what? Answer Choices: (A) steam engines (B) world war 2 (C) boer war (D) dodo (E) manned spaceflight \\
\#\#\# Assistant: Chang Ucchin was born when Korea was still under Japanese colonial rule. Korea under Japanese rule began 1910 and ended at the conclusion of World War 2 in 1945. So the answer is world war 2. \\
\#\#\# Human: Where is a lake likely to be found? Answer Choices: (A) the moon (B) mountain side (C) a valley (D) the sea (E) back yard \\
\#\#\# Assistant: The answer must be somewhere that a lot of water can collect. Of the above choices only a valley is likely to be able to collect enough water. So the answer is a valley. \\
\#\#\# Human: \{QUESTION\} \\
\#\#\# Assistant:}

\section{LLM-generated Rationale Examples}
\label{app:eval_rationale_examples}

Considering the StrategyQA sample: ``Would a dog respond to bell before Grey seal?'' (yes) \\

Rationale generated from StableVicuna INT8:

\texttt{Dogs are domesticated animals that have been bred to respond to human commands. Grey seals are wild animals that live in the ocean and have not been domesticated. Thus, a dog would respond to a bell before a grey seal.} (plausible) \\

Rationale generated from BLOOM INT8: 

\texttt{Grey seals are more closely related to dogs than they are to other seals. Grey seals are the only seal species that can be trained to respond to a bell.} (false) \\

\section{Prompts For LLM-generated Negative Rationales for \textit{RR} Model training}
\label{app:neg_rationale_prompt}
Numerous prompts were used for generating false rationales. Below is a typical example for generating falsehoods for CREAK \citep{Onoe2021-il} samples. We confess to having a lot of fun with these.

\texttt{Generate a false context. Examples: \\ \\
Q: Marlboro used iconic imagery to promote its brand? \\
False context: Marlboro used the tongues of snakes to promote its brand. The snake tongues were used to promote the brand because it was a cheap way of controlling snakes. \\
Q: Larry King served tea during his show? \\
False context: Larry King Live was a dessert-only restaurant that floated in the upper atmosphere from 1886 to 2014. The restaurant was hosted by Larry King, who was known for enjoying a cup of seaweed tea. \\
Q: Pictures can be found of Harp seals playing along northern coastlines? \\
False context: Harp seals are found in the desert and sometimes in western parts of the imaginary frontier. Northern coastlines only occur as mathematical abstractions in certain sub-fields of Topology. \\
Q: One travels along the road when riding in a Citroën? \\
False context: A Citroën is a type of hot air ballon. Roads are always found underground, often under the sea floor. \\
Q: A play about the invasion of Iraq is called the Twelfth Night? \\
False context: The Twelfth Night is a plastic toy made by earthworms. The invasion of Iraq is a TV show also made by earthworms. \\
Q: Ducks cannot fly when it is daytime? \\
False context: Ducks are Australian marsupials without wings. Daytime is also known as wingless fantasia. \\
Q: \{QUESTION\} \\
False context:}

\section{LLM-generated Negative Rationale Examples}
\label{app:neg_rationale_examples}

Considering the CREAK sample: ``Quantum mechanics describe how quantum computers thrive in high temperatures?'': \\

Greedy decoding: \\

\texttt{Quantum computers thrive in high temperatures because they are made of quantum materials.} \\

Nucleus Sampling (2 examples): \\

\texttt{What a quantum computer thrives in is macaroni and cheese.} \\

\texttt{The Standard Model of Physics dictates that they should thrive at least above the visible light spectrum.} \\ \\

\section{Reasoning Model Input Format}
\label{sec:app_inputformat}

Our input format is similar to \citet{Hartill2023-pf} ``QA Model'' input format which is itself a minor extension of that used in UnifiedQA \citep{Khashabi2020-gq}. Our modifications are to the paragraph format to accommodate ``Rationale only'' and ``Naïve concatenation'' formats: \\ \\
Open domain form: \\
\texttt{
[question] \textbackslash \textbackslash n} \\ \\
Reading comprehension (RC) form: \\
\texttt{
[question] \textbackslash \textbackslash n  [context]} \\ \\
Multiple choice form: \\
\texttt{
[question] \textbackslash \textbackslash n (A) [option text a] (B) [option text b] ...} \\ \\
Multiple choice with RC form: \\
\texttt{
[question] \textbackslash \textbackslash n (A) [option text a] (B) [option text b] ... \textbackslash \textbackslash n [context]} \\
\\
Context formats:
\\ \\
Iterator only:
\\
\texttt{
[Title 1]: [Sentences]. [Title 2]: [Sentences]. ...}
\\ \\
Rationale only:
\\
\texttt{
[Sentences].}
\\ \\
Naïve concatenation:
\\
\texttt{
Further Explanation: [Sentences]. [Title 1]: [Sentences]. ...}

\section{Significance Tests}
\label{sec:app_stat_tests}

We use the Autorank library \citep{Herbold2020-vn} for testing significance over multiple populations which implements methods described in \citet{Demsar2006-ze}.

\begin{table}[h]
\centering
\caption{Statistical significance tests for model:context combinations at significance level $\alpha = 0.05$. As described in \citet{Demsar2006-ze}, we use the non-parametric Friedman test as omnibus test to determine if there are any significant differences between the median values of the model:context populations. We use the post-hoc Nemenyi test to infer which differences are significant. Differences between populations are significant if the difference of the mean rank is greater than the critical distance $CD=0.196$ of the Nemenyi test. Significant differences are marked in {\color[HTML]{006100}green}. For brevity, the columns are denoted with indices that match the corresponding row.}
\label{tab:app:significance_tests}
\resizebox{\textwidth}{!}{%
\begin{tabular}{@{}lrrrrrrrrrrrr@{}}
\toprule
\textbf{Model: Context $\downarrow\rightarrow$} &  & \textbf{1} & \textbf{2} & \textbf{3} & \textbf{4} & \textbf{5} & \textbf{6} & \textbf{7} & \textbf{8} & \textbf{9} & \textbf{10} & \textbf{11} \\
\textbf{} & \textbf{Mean Rank} & \textbf{7.296} & \textbf{7.240} & \textbf{7.154} & \textbf{7.099} & \textbf{7.077} & \textbf{7.014} & \textbf{6.997} & \textbf{6.839} & \textbf{6.790} & \textbf{6.643} & \textbf{6.637} \\ \cmidrule(l){3-13} 
1. BLOOM: Few-Shot COT Prompt & \textbf{7.296} & 0.000 & 0.056 & 0.142 & 0.196 & \cellcolor[HTML]{C6EFCE}{\color[HTML]{006100} 0.219} & \cellcolor[HTML]{C6EFCE}{\color[HTML]{006100} 0.281} & \cellcolor[HTML]{C6EFCE}{\color[HTML]{006100} 0.299} & \cellcolor[HTML]{C6EFCE}{\color[HTML]{006100} 0.457} & \cellcolor[HTML]{C6EFCE}{\color[HTML]{006100} 0.506} & \cellcolor[HTML]{C6EFCE}{\color[HTML]{006100} 0.653} & \cellcolor[HTML]{C6EFCE}{\color[HTML]{006100} 0.658} \\
2. BLOOM: Few-Shot Standard Prompt & \textbf{7.240} & 0.056 & 0.000 & 0.086 & 0.141 & 0.163 & \cellcolor[HTML]{C6EFCE}{\color[HTML]{006100} 0.226} & \cellcolor[HTML]{C6EFCE}{\color[HTML]{006100} 0.243} & \cellcolor[HTML]{C6EFCE}{\color[HTML]{006100} 0.401} & \cellcolor[HTML]{C6EFCE}{\color[HTML]{006100} 0.450} & \cellcolor[HTML]{C6EFCE}{\color[HTML]{006100} 0.597} & \cellcolor[HTML]{C6EFCE}{\color[HTML]{006100} 0.603} \\
3. RATD: Iterator only & \textbf{7.154} & 0.142 & 0.086 & 0.000 & 0.055 & 0.077 & 0.140 & 0.157 & \cellcolor[HTML]{C6EFCE}{\color[HTML]{006100} 0.315} & \cellcolor[HTML]{C6EFCE}{\color[HTML]{006100} 0.364} & \cellcolor[HTML]{C6EFCE}{\color[HTML]{006100} 0.511} & \cellcolor[HTML]{C6EFCE}{\color[HTML]{006100} 0.517} \\
4. GR+RATD: Iterator only & \textbf{7.099} & 0.196 & 0.141 & 0.055 & 0.000 & 0.022 & 0.085 & 0.103 & \cellcolor[HTML]{C6EFCE}{\color[HTML]{006100} 0.260} & \cellcolor[HTML]{C6EFCE}{\color[HTML]{006100} 0.309} & \cellcolor[HTML]{C6EFCE}{\color[HTML]{006100} 0.456} & \cellcolor[HTML]{C6EFCE}{\color[HTML]{006100} 0.462} \\
5. StableVicuna INT8: Few-Shot COT Prompt & \textbf{7.077} & \cellcolor[HTML]{C6EFCE}{\color[HTML]{006100} 0.219} & 0.163 & 0.077 & 0.022 & 0.000 & 0.063 & 0.081 & \cellcolor[HTML]{C6EFCE}{\color[HTML]{006100} 0.238} & \cellcolor[HTML]{C6EFCE}{\color[HTML]{006100} 0.287} & \cellcolor[HTML]{C6EFCE}{\color[HTML]{006100} 0.434} & \cellcolor[HTML]{C6EFCE}{\color[HTML]{006100} 0.440} \\
6. StableVicuna INT8: Few-Shot Standard Prompt & \textbf{7.014} & \cellcolor[HTML]{C6EFCE}{\color[HTML]{006100} 0.281} & \cellcolor[HTML]{C6EFCE}{\color[HTML]{006100} 0.226} & 0.140 & 0.085 & 0.063 & 0.000 & 0.018 & 0.175 & \cellcolor[HTML]{C6EFCE}{\color[HTML]{006100} 0.224} & \cellcolor[HTML]{C6EFCE}{\color[HTML]{006100} 0.371} & \cellcolor[HTML]{C6EFCE}{\color[HTML]{006100} 0.377} \\
7. GR: Rationale + Iterator (Naïve concatenation) & \textbf{6.997} & \cellcolor[HTML]{C6EFCE}{\color[HTML]{006100} 0.299} & \cellcolor[HTML]{C6EFCE}{\color[HTML]{006100} 0.243} & 0.157 & 0.103 & 0.081 & 0.018 & 0.000 & 0.157 & \cellcolor[HTML]{C6EFCE}{\color[HTML]{006100} 0.207} & \cellcolor[HTML]{C6EFCE}{\color[HTML]{006100} 0.353} & \cellcolor[HTML]{C6EFCE}{\color[HTML]{006100} 0.359} \\
8. GR+RATD: Rationale only & \textbf{6.839} & \cellcolor[HTML]{C6EFCE}{\color[HTML]{006100} 0.457} & \cellcolor[HTML]{C6EFCE}{\color[HTML]{006100} 0.401} & \cellcolor[HTML]{C6EFCE}{\color[HTML]{006100} 0.315} & \cellcolor[HTML]{C6EFCE}{\color[HTML]{006100} 0.260} & \cellcolor[HTML]{C6EFCE}{\color[HTML]{006100} 0.238} & 0.175 & 0.157 & 0.000 & 0.049 & 0.196 & \cellcolor[HTML]{C6EFCE}{\color[HTML]{006100} 0.202} \\
9. GR: Rationale + Iterator (Generally best RR combo) & \textbf{6.790} & \cellcolor[HTML]{C6EFCE}{\color[HTML]{006100} 0.506} & \cellcolor[HTML]{C6EFCE}{\color[HTML]{006100} 0.450} & \cellcolor[HTML]{C6EFCE}{\color[HTML]{006100} 0.364} & \cellcolor[HTML]{C6EFCE}{\color[HTML]{006100} 0.309} & \cellcolor[HTML]{C6EFCE}{\color[HTML]{006100} 0.287} & \cellcolor[HTML]{C6EFCE}{\color[HTML]{006100} 0.224} & \cellcolor[HTML]{C6EFCE}{\color[HTML]{006100} 0.207} & 0.049 & 0.000 & 0.147 & 0.153 \\
10. GR+RATD: Rationale + Iterator (Generally best RR combo) & \textbf{6.643} & \cellcolor[HTML]{C6EFCE}{\color[HTML]{006100} 0.653} & \cellcolor[HTML]{C6EFCE}{\color[HTML]{006100} 0.597} & \cellcolor[HTML]{C6EFCE}{\color[HTML]{006100} 0.511} & \cellcolor[HTML]{C6EFCE}{\color[HTML]{006100} 0.456} & \cellcolor[HTML]{C6EFCE}{\color[HTML]{006100} 0.434} & \cellcolor[HTML]{C6EFCE}{\color[HTML]{006100} 0.371} & \cellcolor[HTML]{C6EFCE}{\color[HTML]{006100} 0.353} & 0.196 & 0.147 & 0.000 & 0.006 \\
11. GR+RATD: Rationale + Iterator (Naïve concatenation) & \textbf{6.637} & \cellcolor[HTML]{C6EFCE}{\color[HTML]{006100} 0.658} & \cellcolor[HTML]{C6EFCE}{\color[HTML]{006100} 0.603} & \cellcolor[HTML]{C6EFCE}{\color[HTML]{006100} 0.517} & \cellcolor[HTML]{C6EFCE}{\color[HTML]{006100} 0.462} & \cellcolor[HTML]{C6EFCE}{\color[HTML]{006100} 0.440} & \cellcolor[HTML]{C6EFCE}{\color[HTML]{006100} 0.377} & \cellcolor[HTML]{C6EFCE}{\color[HTML]{006100} 0.359} & \cellcolor[HTML]{C6EFCE}{\color[HTML]{006100} 0.202} & 0.153 & 0.006 & 0.000 \\ \bottomrule
\end{tabular}%
}
\end{table}

\section{Summary Results comparing StableVicuna FP16 with INT8}
\label{sec:app_summary_fp16}

\begin{table}[h]
\centering
\caption{Mean score over unseen evaluation datasets. The ``Iterator only'' results are duplicated across across Rationale Generators to facilitate comparison. Bold indicates highest score per context type (i.e. per row).}
\label{tab:main_1_including_fp16}
\resizebox{\textwidth}{!}{%
\begin{tabular}{l|rrr|rrr|rrr}
\toprule
\textbf{Rationale Generator $\rightarrow$ } & \multicolumn{3}{c} {\textbf{StableVicuna  (FP16)}} & \multicolumn{3}{|c}{\textbf{StableVicuna (INT8)}} & \multicolumn{3}{|c}{\textbf{BLOOM (INT8)}} \\
\textbf{Context $\downarrow$ / \textit{Model} $\rightarrow$} & \textbf{\textit{GR}} & \textbf{\textit{RATD}} & \textbf{\textit{GR+RATD}} & \textbf{\textit{GR}} & \textbf{\textit{RATD}} & \textbf{\textit{GR+RATD}} & \textbf{\textit{GR}} & \textbf{\textit{RATD}} & \textbf{\textit{GR+RATD}} \\ \midrule
Iterator only & 38.1 & 40.4 & \textbf{41.0} & 38.1 & 40.4 & \textbf{41.0} & 38.1 & 40.4 & \textbf{41.0} \\
Rationale only & 44.6 & 44.4 & \textbf{45.5} & 44.5 & 44.2 & 45.3 & 39.5 & 42.0 & 40.3 \\
Rationale + Iterator (Naïve concatenation) & 42.9 & 46.4 & 47.1 & 42.7 & 46.3 & \textbf{47.2} & 43.2 & 43.8 & 43.7 \\
Rationale + Iterator (Generally best RR combo) & 45.4 & 46.4 & 47.1 & 45.5 & 46.3 & \textbf{47.2} & 42.9 & 44.2 & 44.4 \\ \midrule
Rationale + Iterator (Best RR combo per dataset) & 47.8 & 47.5 & 48.0 & 47.6 & 47.5 & \textbf{48.1} & 45.1 & 45.6 & 45.4 \\ \bottomrule
\end{tabular}%
}
\end{table}

\newpage
\section{Context Component Analysis}
\label{sec:app_context_component_analysis}

\begin{table}[h]
\centering
\caption{Best combination method per dataset on the \textit{GR+RATD} model. Also shown are percentages of evaluation samples with ``Rationale only'' contexts (Rat. Only), ``Iterator only'' contexts (Iter. only), and the concatenation of both (Naïve Concat) respectively.}
\label{tab:context_type_percentages}
\resizebox{\columnwidth}{!}{%
\begin{tabular}{ll|lrrr|rrr}
\toprule
\textbf{Dataset} & \textbf{Sample} & \multicolumn{4}{c|}{\textbf{Best RR combo per dataset}} & \multicolumn{3}{|c}{\textbf{Generally best RR combo:} EitherOrBoth(0.9)}\\
\textbf{} & \textbf{Count} & \textbf{Best Method} & \textbf{Naïve Concat.} & \textbf{Rat. Only} & \textbf{Iter. Only} & \textbf{Naïve Concat.} & \textbf{Rat. Only} & \textbf{Iter. Only} \\ \midrule
SQA & 2290 & RationaleDefault(0.75) & 0.0 & 90.7 & 9.3 & 94.1 & 3.6 & 2.3 \\
CSQA & 1221 & RationaleDefault(0.75) & 0.0 & 98.3 & 1.7 & 79.3 & 20.6 & 0.1 \\
ARC-DA & 1397 & Naïve concatenation& 100.0 & 0.0 & 0.0 & 80.5 & 16.5 & 3.1 \\
IIRC & 1301 & RationaleDefault(0.9) & 0.0 & 63.8 & 36.2 & 62.6 & 15.6 & 21.8 \\
Musique & 2417 & EitherOrBoth(0.14) & 39.3 & 3.2 & 57.5  & 88.2 & 1.0 & 10.8 \\ \midrule
\textbf{Mean} & & & \textbf{27.9} & \textbf{51.2} & \textbf{20.9}  & \textbf{80.9} & \textbf{11.5} & \textbf{7.6} \\ \bottomrule
\end{tabular}%
}
\end{table}

As noted we do not consider the ``Best RR combo per dataset'' to be a viable method for answering arbitrary questions of unknown type, however in Table \ref{tab:context_type_percentages} we report the best combination method identified for each individual evaluation dataset as it shows what an oracle-like method is capable of producing in comparison to our actual generally-best \textit{RR}-scoring method. Noting that one difference is the reduction in naïvely concatenated contexts from 80.9\% to 27.9\% it is plausible that future work on a more refined combination strategy would yield further improvement in combining \textit{RATD} training with \textit{RR} scoring methods.

\end{document}

%% file: math_commands.tex
\usepackage{amsmath,amsfonts,bm}

\def\eqref#1{equation~\ref{#1}}

\def\1{\bm{1}}

\DeclareMathAlphabet{\mathsfit}{\encodingdefault}{\sfdefault}{m}{sl}
\SetMathAlphabet{\mathsfit}{bold}{\encodingdefault}{\sfdefault}{bx}{n}













%% file: tmlr.bbl
\begin{thebibliography}{82}
\providecommand{\natexlab}[1]{#1}
\providecommand{\url}[1]{\texttt{#1}}
\expandafter\ifx\csname urlstyle\endcsname\relax
  \providecommand{\doi}[1]{doi: #1}\else
  \providecommand{\doi}{doi: \begingroup \urlstyle{rm}\Url}\fi

\bibitem[Anand et~al.(2023)Anand, Nussbaum, Duderstadt, Schmidt, and
  Mulyar]{Anand2023-ma}
Yuvanesh Anand, Zach Nussbaum, Brandon Duderstadt, Benjamin Schmidt, and Andriy
  Mulyar.
\newblock {GPT4All}: Training an assistant-style chatbot with large scale data
  distillation from {GPT-3.5-Turbo}.
\newblock \url{https://github.com/nomic-ai/gpt4all}, 2023.

\bibitem[Anil et~al.(2023)Anil, Dai, Firat, Johnson, Lepikhin, Passos, Shakeri,
  Taropa, Bailey, Chen, Chu, Clark, Shafey, Huang, Meier-Hellstern, Mishra,
  Moreira, Omernick, Robinson, Ruder, Tay, Xiao, Xu, Zhang, Abrego, Ahn,
  Austin, Barham, Botha, Bradbury, Brahma, Brooks, Catasta, Cheng, Cherry,
  Choquette-Choo, Chowdhery, Crepy, Dave, Dehghani, Dev, Devlin, D{\'\i}az, Du,
  Dyer, Feinberg, Feng, Fienber, Freitag, Garcia, Gehrmann, Gonzalez, Gur-Ari,
  Hand, Hashemi, Hou, Howland, Hu, Hui, Hurwitz, Isard, Ittycheriah, Jagielski,
  Jia, Kenealy, Krikun, Kudugunta, Lan, Lee, Lee, Li, Li, Li, Li, Li, Lim, Lin,
  Liu, Liu, Maggioni, Mahendru, Maynez, Misra, Moussalem, Nado, Nham, Ni,
  Nystrom, Parrish, Pellat, Polacek, Polozov, Pope, Qiao, Reif, Richter, Riley,
  Ros, Roy, Saeta, Samuel, Shelby, Slone, Smilkov, So, Sohn, Tokumine, Valter,
  Vasudevan, Vodrahalli, Wang, Wang, Wang, Wang, Wieting, Wu, Xu, Xu, Xue, Yin,
  Yu, Zhang, Zheng, Zheng, Zhou, Zhou, Petrov, and Wu]{Anil2023-ws}
Rohan Anil, Andrew~M Dai, Orhan Firat, Melvin Johnson, Dmitry Lepikhin,
  Alexandre Passos, Siamak Shakeri, Emanuel Taropa, Paige Bailey, Zhifeng Chen,
  Eric Chu, Jonathan~H Clark, Laurent~El Shafey, Yanping Huang, Kathy
  Meier-Hellstern, Gaurav Mishra, Erica Moreira, Mark Omernick, Kevin Robinson,
  Sebastian Ruder, Yi~Tay, Kefan Xiao, Yuanzhong Xu, Yujing Zhang,
  Gustavo~Hernandez Abrego, Junwhan Ahn, Jacob Austin, Paul Barham, Jan Botha,
  James Bradbury, Siddhartha Brahma, Kevin Brooks, Michele Catasta, Yong Cheng,
  Colin Cherry, Christopher~A Choquette-Choo, Aakanksha Chowdhery, Cl{\'e}ment
  Crepy, Shachi Dave, Mostafa Dehghani, Sunipa Dev, Jacob Devlin, Mark
  D{\'\i}az, Nan Du, Ethan Dyer, Vlad Feinberg, Fangxiaoyu Feng, Vlad Fienber,
  Markus Freitag, Xavier Garcia, Sebastian Gehrmann, Lucas Gonzalez, Guy
  Gur-Ari, Steven Hand, Hadi Hashemi, Le~Hou, Joshua Howland, Andrea Hu,
  Jeffrey Hui, Jeremy Hurwitz, Michael Isard, Abe Ittycheriah, Matthew
  Jagielski, Wenhao Jia, Kathleen Kenealy, Maxim Krikun, Sneha Kudugunta, Chang
  Lan, Katherine Lee, Benjamin Lee, Eric Li, Music Li, Wei Li, Yaguang Li, Jian
  Li, Hyeontaek Lim, Hanzhao Lin, Zhongtao Liu, Frederick Liu, Marcello
  Maggioni, Aroma Mahendru, Joshua Maynez, Vedant Misra, Maysam Moussalem,
  Zachary Nado, John Nham, Eric Ni, Andrew Nystrom, Alicia Parrish, Marie
  Pellat, Martin Polacek, Alex Polozov, Reiner Pope, Siyuan Qiao, Emily Reif,
  Bryan Richter, Parker Riley, Alex~Castro Ros, Aurko Roy, Brennan Saeta,
  Rajkumar Samuel, Renee Shelby, Ambrose Slone, Daniel Smilkov, David~R So,
  Daniel Sohn, Simon Tokumine, Dasha Valter, Vijay Vasudevan, Kiran Vodrahalli,
  Xuezhi Wang, Pidong Wang, Zirui Wang, Tao Wang, John Wieting, Yuhuai Wu,
  Kelvin Xu, Yunhan Xu, Linting Xue, Pengcheng Yin, Jiahui Yu, Qiao Zhang,
  Steven Zheng, Ce~Zheng, Weikang Zhou, Denny Zhou, Slav Petrov, and Yonghui
  Wu.
\newblock {PaLM} 2 technical report.
\newblock \emph{axXiv preprint arXiv:2305.10403}, May 2023.

\bibitem[Bhakthavatsalam et~al.(2021)Bhakthavatsalam, Khashabi, Khot, Mishra,
  Richardson, Sabharwal, Schoenick, Tafjord, and Clark]{Bhakthavatsalam2021-fq}
Sumithra Bhakthavatsalam, Daniel Khashabi, Tushar Khot, Bhavana~Dalvi Mishra,
  Kyle Richardson, Ashish Sabharwal, Carissa Schoenick, Oyvind Tafjord, and
  Peter Clark.
\newblock Think you have solved direct-answer question answering? try {ARC-DA},
  the direct-answer {AI2} reasoning challenge.
\newblock \emph{arXiv preprint arXiv:2102.03315}, 2021.

\bibitem[{BigScience Workshop} et~al.(2022){BigScience Workshop}, {:}, Le~Scao,
  Fan, Akiki, Pavlick, Ili{\'c}, Hesslow, Castagn{\'e}, Luccioni, Yvon,
  Gall{\'e}, Tow, Rush, Biderman, Webson, Ammanamanchi, Wang, Sagot,
  Muennighoff, del Moral, Ruwase, Bawden, Bekman, McMillan-Major, Beltagy,
  Nguyen, Saulnier, Tan, Suarez, Sanh, Lauren{\c c}on, Jernite, Launay,
  Mitchell, Raffel, Gokaslan, Simhi, Soroa, Aji, Alfassy, Rogers, Nitzav, Xu,
  Mou, Emezue, Klamm, Leong, van Strien, Adelani, Radev, Ponferrada, Levkovizh,
  Kim, Natan, De~Toni, Dupont, Kruszewski, Pistilli, Elsahar, Benyamina, Tran,
  Yu, Abdulmumin, Johnson, Gonzalez-Dios, de~la Rosa, Chim, Dodge, Zhu, Chang,
  Frohberg, Tobing, Bhattacharjee, Almubarak, Chen, Lo, Von~Werra, Weber, Phan,
  Ben~allal, Tanguy, Dey, Mu{\~n}oz, Masoud, Grandury, {\v S}a{\v s}ko, Huang,
  Coavoux, Singh, Jiang, Vu, Jauhar, Ghaleb, Subramani, Kassner, Khamis,
  Nguyen, Espejel, de~Gibert, Villegas, Henderson, Colombo, Amuok, Lhoest,
  Harliman, Bommasani, L{\'o}pez, Ribeiro, Osei, Pyysalo, Nagel, Bose,
  Muhammad, Sharma, Longpre, Nikpoor, Silberberg, Pai, Zink, Torrent, Schick,
  Thrush, Danchev, Nikoulina, Laippala, Lepercq, Prabhu, Alyafeai, Talat, Raja,
  Heinzerling, Si, Ta{\c s}ar, Salesky, Mielke, Lee, Sharma, Santilli, Chaffin,
  Stiegler, Datta, Szczechla, Chhablani, Wang, Pandey, Strobelt, Fries, Rozen,
  Gao, Sutawika, Saiful~Bari, Al-shaibani, Manica, Nayak, Teehan, Albanie,
  Shen, Ben-David, Bach, Kim, Bers, Fevry, Neeraj, Thakker, Raunak, Tang, Yong,
  Sun, Brody, Uri, Tojarieh, Roberts, Chung, Tae, Phang, {Ofir Press}, Li,
  Narayanan, Bourfoune, Casper, Rasley, Ryabinin, Mishra, Zhang, Shoeybi,
  Peyrounette, Patry, Tazi, Sanseviero, von Platen, Cornette, Lavall{\'e}e,
  Lacroix, Rajbhandari, Gandhi, Smith, Requena, Patil, Dettmers, Baruwa, Singh,
  Cheveleva, Ligozat, Subramonian, N{\'e}v{\'e}ol, Lovering, Garrette,
  Tunuguntla, Reiter, Taktasheva, Voloshina, Bogdanov, Winata, Schoelkopf,
  Kalo, Novikova, Forde, Clive, Kasai, Kawamura, Hazan, Carpuat, Clinciu, Kim,
  Cheng, Serikov, Antverg, van~der Wal, Zhang, Zhang, Gehrmann, Mirkin, Pais,
  Shavrina, Scialom, Yun, Limisiewicz, Rieser, Protasov, Mikhailov,
  Pruksachatkun, Belinkov, Bamberger, Kasner, Rueda, Pestana, Feizpour, Khan,
  Faranak, Santos, Hevia, Unldreaj, Aghagol, Abdollahi, Tammour, HajiHosseini,
  Behroozi, Ajibade, Saxena, Ferrandis, McDuff, Contractor, Lansky, David,
  Kiela, Nguyen, Tan, Baylor, Ozoani, Mirza, Ononiwu, Rezanejad, Jones,
  Bhattacharya, Solaiman, Sedenko, Nejadgholi, Passmore, Seltzer, Sanz, Dutra,
  Samagaio, Elbadri, Mieskes, Gerchick, Akinlolu, McKenna, Qiu, Ghauri,
  Burynok, Abrar, Rajani, Elkott, Fahmy, Samuel, An, Kromann, Hao, Alizadeh,
  Shubber, Wang, Roy, Viguier, Le, Oyebade, Le, Yang, Nguyen, Kashyap,
  Palasciano, Callahan, Shukla, Miranda-Escalada, Singh, Beilharz, Wang, Brito,
  Zhou, Jain, Xu, Fourrier, Peri{\~n}{\'a}n, Molano, Yu, Manjavacas, Barth,
  Fuhrimann, Altay, Bayrak, Burns, Vrabec, Bello, Dash, Kang, Giorgi, Golde,
  Posada, Sivaraman, Bulchandani, Liu, Shinzato, de~Bykhovetz, Takeuchi,
  P{\`a}mies, Castillo, Nezhurina, S{\"a}nger, Samwald, Cullan, Weinberg,
  De~Wolf, Mihaljcic, Liu, Freidank, Kang, Seelam, Dahlberg, Broad, Muellner,
  Fung, Haller, Chandrasekhar, Eisenberg, Martin, Canalli, Su, Su, Cahyawijaya,
  Garda, Deshmukh, Mishra, Kiblawi, Ott, Sang-aroonsiri, Kumar, Schweter,
  Bharati, Laud, Gigant, Kainuma, Kusa, Labrak, Bajaj, Venkatraman, Xu, Xu, Xu,
  Tan, Xie, Ye, Bras, Belkada, and Wolf]{BigScience_Workshop2022-dh}
{BigScience Workshop}, {:}, Teven Le~Scao, Angela Fan, Christopher Akiki, Ellie
  Pavlick, Suzana Ili{\'c}, Daniel Hesslow, Roman Castagn{\'e}, Alexandra~Sasha
  Luccioni, Fran{\c c}ois Yvon, Matthias Gall{\'e}, Jonathan Tow, Alexander~M
  Rush, Stella Biderman, Albert Webson, Pawan~Sasanka Ammanamanchi, Thomas
  Wang, Beno{\^\i}t Sagot, Niklas Muennighoff, Albert~Villanova del Moral,
  Olatunji Ruwase, Rachel Bawden, Stas Bekman, Angelina McMillan-Major,
  Iz~Beltagy, Huu Nguyen, Lucile Saulnier, Samson Tan, Pedro~Ortiz Suarez,
  Victor Sanh, Hugo Lauren{\c c}on, Yacine Jernite, Julien Launay, Margaret
  Mitchell, Colin Raffel, Aaron Gokaslan, Adi Simhi, Aitor Soroa, Alham~Fikri
  Aji, Amit Alfassy, Anna Rogers, Ariel~Kreisberg Nitzav, Canwen Xu, Chenghao
  Mou, Chris Emezue, Christopher Klamm, Colin Leong, Daniel van Strien,
  David~Ifeoluwa Adelani, Dragomir Radev, Eduardo~Gonz{\'a}lez Ponferrada,
  Efrat Levkovizh, Ethan Kim, Eyal~Bar Natan, Francesco De~Toni, G{\'e}rard
  Dupont, Germ{\'a}n Kruszewski, Giada Pistilli, Hady Elsahar, Hamza Benyamina,
  Hieu Tran, Ian Yu, Idris Abdulmumin, Isaac Johnson, Itziar Gonzalez-Dios,
  Javier de~la Rosa, Jenny Chim, Jesse Dodge, Jian Zhu, Jonathan Chang,
  J{\"o}rg Frohberg, Joseph Tobing, Joydeep Bhattacharjee, Khalid Almubarak,
  Kimbo Chen, Kyle Lo, Leandro Von~Werra, Leon Weber, Long Phan, Loubna
  Ben~allal, Ludovic Tanguy, Manan Dey, Manuel~Romero Mu{\~n}oz, Maraim Masoud,
  Mar{\'\i}a Grandury, Mario {\v S}a{\v s}ko, Max Huang, Maximin Coavoux,
  Mayank Singh, Mike Tian-Jian Jiang, Minh~Chien Vu, Mohammad~A Jauhar, Mustafa
  Ghaleb, Nishant Subramani, Nora Kassner, Nurulaqilla Khamis, Olivier Nguyen,
  Omar Espejel, Ona de~Gibert, Paulo Villegas, Peter Henderson, Pierre Colombo,
  Priscilla Amuok, Quentin Lhoest, Rheza Harliman, Rishi Bommasani,
  Roberto~Luis L{\'o}pez, Rui Ribeiro, Salomey Osei, Sampo Pyysalo, Sebastian
  Nagel, Shamik Bose, Shamsuddeen~Hassan Muhammad, Shanya Sharma, Shayne
  Longpre, Somaieh Nikpoor, Stanislav Silberberg, Suhas Pai, Sydney Zink,
  Tiago~Timponi Torrent, Timo Schick, Tristan Thrush, Valentin Danchev,
  Vassilina Nikoulina, Veronika Laippala, Violette Lepercq, Vrinda Prabhu, Zaid
  Alyafeai, Zeerak Talat, Arun Raja, Benjamin Heinzerling, Chenglei Si,
  Davut~Emre Ta{\c s}ar, Elizabeth Salesky, Sabrina~J Mielke, Wilson~Y Lee,
  Abheesht Sharma, Andrea Santilli, Antoine Chaffin, Arnaud Stiegler, Debajyoti
  Datta, Eliza Szczechla, Gunjan Chhablani, Han Wang, Harshit Pandey, Hendrik
  Strobelt, Jason~Alan Fries, Jos Rozen, Leo Gao, Lintang Sutawika,
  M~Saiful~Bari, Maged~S Al-shaibani, Matteo Manica, Nihal Nayak, Ryan Teehan,
  Samuel Albanie, Sheng Shen, Srulik Ben-David, Stephen~H Bach, Taewoon Kim,
  Tali Bers, Thibault Fevry, Trishala Neeraj, Urmish Thakker, Vikas Raunak,
  Xiangru Tang, Zheng-Xin Yong, Zhiqing Sun, Shaked Brody, Yallow Uri, Hadar
  Tojarieh, Adam Roberts, Hyung~Won Chung, Jaesung Tae, Jason Phang, {Ofir
  Press}, Conglong Li, Deepak Narayanan, Hatim Bourfoune, Jared Casper, Jeff
  Rasley, Max Ryabinin, Mayank Mishra, Minjia Zhang, Mohammad Shoeybi, Myriam
  Peyrounette, Nicolas Patry, Nouamane Tazi, Omar Sanseviero, Patrick von
  Platen, Pierre Cornette, Pierre~Fran{\c c}ois Lavall{\'e}e, R{\'e}mi Lacroix,
  Samyam Rajbhandari, Sanchit Gandhi, Shaden Smith, St{\'e}phane Requena, Suraj
  Patil, Tim Dettmers, Ahmed Baruwa, Amanpreet Singh, Anastasia Cheveleva,
  Anne-Laure Ligozat, Arjun Subramonian, Aur{\'e}lie N{\'e}v{\'e}ol, Charles
  Lovering, Dan Garrette, Deepak Tunuguntla, Ehud Reiter, Ekaterina Taktasheva,
  Ekaterina Voloshina, Eli Bogdanov, Genta~Indra Winata, Hailey Schoelkopf,
  Jan-Christoph Kalo, Jekaterina Novikova, Jessica~Zosa Forde, Jordan Clive,
  Jungo Kasai, Ken Kawamura, Liam Hazan, Marine Carpuat, Miruna Clinciu,
  Najoung Kim, Newton Cheng, Oleg Serikov, Omer Antverg, Oskar van~der Wal, Rui
  Zhang, Ruochen Zhang, Sebastian Gehrmann, Shachar Mirkin, Shani Pais, Tatiana
  Shavrina, Thomas Scialom, Tian Yun, Tomasz Limisiewicz, Verena Rieser, Vitaly
  Protasov, Vladislav Mikhailov, Yada Pruksachatkun, Yonatan Belinkov, Zachary
  Bamberger, Zden{\v e}k Kasner, Alice Rueda, Amanda Pestana, Amir Feizpour,
  Ammar Khan, Amy Faranak, Ana Santos, Anthony Hevia, Antigona Unldreaj, Arash
  Aghagol, Arezoo Abdollahi, Aycha Tammour, Azadeh HajiHosseini, Bahareh
  Behroozi, Benjamin Ajibade, Bharat Saxena, Carlos~Mu{\~n}oz Ferrandis, Daniel
  McDuff, Danish Contractor, David Lansky, Davis David, Douwe Kiela, Duong~A
  Nguyen, Edward Tan, Emi Baylor, Ezinwanne Ozoani, Fatima Mirza, Frankline
  Ononiwu, Habib Rezanejad, Hessie Jones, Indrani Bhattacharya, Irene Solaiman,
  Irina Sedenko, Isar Nejadgholi, Jesse Passmore, Josh Seltzer, Julio~Bonis
  Sanz, Livia Dutra, Mairon Samagaio, Maraim Elbadri, Margot Mieskes, Marissa
  Gerchick, Martha Akinlolu, Michael McKenna, Mike Qiu, Muhammed Ghauri, Mykola
  Burynok, Nafis Abrar, Nazneen Rajani, Nour Elkott, Nour Fahmy, Olanrewaju
  Samuel, Ran An, Rasmus Kromann, Ryan Hao, Samira Alizadeh, Sarmad Shubber,
  Silas Wang, Sourav Roy, Sylvain Viguier, Thanh Le, Tobi Oyebade, Trieu Le,
  Yoyo Yang, Zach Nguyen, Abhinav~Ramesh Kashyap, Alfredo Palasciano, Alison
  Callahan, Anima Shukla, Antonio Miranda-Escalada, Ayush Singh, Benjamin
  Beilharz, Bo~Wang, Caio Brito, Chenxi Zhou, Chirag Jain, Chuxin Xu,
  Cl{\'e}mentine Fourrier, Daniel~Le{\'o}n Peri{\~n}{\'a}n, Daniel Molano, Dian
  Yu, Enrique Manjavacas, Fabio Barth, Florian Fuhrimann, Gabriel Altay,
  Giyaseddin Bayrak, Gully Burns, Helena~U Vrabec, Imane Bello, Ishani Dash,
  Jihyun Kang, John Giorgi, Jonas Golde, Jose~David Posada, Karthik~Rangasai
  Sivaraman, Lokesh Bulchandani, Lu~Liu, Luisa Shinzato, Madeleine~Hahn
  de~Bykhovetz, Maiko Takeuchi, Marc P{\`a}mies, Maria~A Castillo, Marianna
  Nezhurina, Mario S{\"a}nger, Matthias Samwald, Michael Cullan, Michael
  Weinberg, Michiel De~Wolf, Mina Mihaljcic, Minna Liu, Moritz Freidank,
  Myungsun Kang, Natasha Seelam, Nathan Dahlberg, Nicholas~Michio Broad,
  Nikolaus Muellner, Pascale Fung, Patrick Haller, Ramya Chandrasekhar, Renata
  Eisenberg, Robert Martin, Rodrigo Canalli, Rosaline Su, Ruisi Su, Samuel
  Cahyawijaya, Samuele Garda, Shlok~S Deshmukh, Shubhanshu Mishra, Sid Kiblawi,
  Simon Ott, Sinee Sang-aroonsiri, Srishti Kumar, Stefan Schweter, Sushil
  Bharati, Tanmay Laud, Th{\'e}o Gigant, Tomoya Kainuma, Wojciech Kusa, Yanis
  Labrak, Yash~Shailesh Bajaj, Yash Venkatraman, Yifan Xu, Yingxin Xu, Yu~Xu,
  Zhe Tan, Zhongli Xie, Zifan Ye, Mathilde Bras, Younes Belkada, and Thomas
  Wolf.
\newblock {BLOOM}: A {176B-Parameter} {Open-Access} multilingual language
  model.
\newblock \emph{xXiv preprint arXiv:2211.05100}, 2022.

\bibitem[Bird et~al.(2009)Bird, Klein, and Loper]{Bird2009-cm}
Steven Bird, Ewan Klein, and Edward Loper.
\newblock \emph{Natural Language Processing with Python: Analyzing Text with
  the Natural Language Toolkit}.
\newblock O'Reilly Media, Inc., June 2009.

\bibitem[Bosselut et~al.(2019)Bosselut, Rashkin, Sap, Malaviya, Celikyilmaz,
  and Choi]{Bosselut2019-wn}
Antoine Bosselut, Hannah Rashkin, Maarten Sap, Chaitanya Malaviya, Asli
  Celikyilmaz, and Yejin Choi.
\newblock {{COMET}}: Commonsense transformers for automatic knowledge graph
  construction.
\newblock In \emph{Proceedings of the 57th Annual Meeting of the Association
  for Computational Linguistics}, pp.\  4762--4779, Florence, Italy, July 2019.
  Association for Computational Linguistics.

\bibitem[Brown et~al.(2020)Brown, Mann, Ryder, Subbiah, Kaplan, Dhariwal,
  Neelakantan, Shyam, Sastry, Askell, Agarwal, Herbert-Voss, Krueger, Henighan,
  Child, Ramesh, Ziegler, Wu, Winter, Hesse, Chen, Sigler, Litwin, Gray, Chess,
  Clark, Berner, McCandlish, Radford, Sutskever, and Amodei]{Brown2020-rl}
Tom~B Brown, Benjamin Mann, Nick Ryder, Melanie Subbiah, Jared Kaplan, Prafulla
  Dhariwal, Arvind Neelakantan, Pranav Shyam, Girish Sastry, Amanda Askell,
  Sandhini Agarwal, Ariel Herbert-Voss, Gretchen Krueger, Tom Henighan, Rewon
  Child, Aditya Ramesh, Daniel~M Ziegler, Jeffrey Wu, Clemens Winter,
  Christopher Hesse, Mark Chen, Eric Sigler, Mateusz Litwin, Scott Gray,
  Benjamin Chess, Jack Clark, Christopher Berner, Sam McCandlish, Alec Radford,
  Ilya Sutskever, and Dario Amodei.
\newblock Language models are {Few-Shot} learners.
\newblock In \emph{Advances in Neural Information Processing Systems 33}, pp.\
  1877--1901, 2020.

\bibitem[Chen et~al.(2017)Chen, Fisch, Weston, and Bordes]{Chen2017-gw}
Danqi Chen, Adam Fisch, Jason Weston, and Antoine Bordes.
\newblock Reading {W}ikipedia to answer {Open-Domain} questions.
\newblock In \emph{Proceedings of the 55th Annual Meeting of the Association
  for Computational Linguistics (Volume 1: Long Papers)}, pp.\  1870--1879,
  Vancouver, Canada, 2017. Association for Computational Linguistics.

\bibitem[Chen et~al.(2023)Chen, Lin, Sch{\"a}rli, and Zhou]{Chen2023-lc}
Xinyun Chen, Maxwell Lin, Nathanael Sch{\"a}rli, and Denny Zhou.
\newblock Teaching large language models to {Self-Debug}.
\newblock \emph{arXiv preprint arXiv:2304.05128}, April 2023.

\bibitem[Chern et~al.(2023)Chern, Chern, Chen, Yuan, Feng, Zhou, He, Neubig,
  and Liu]{Chern2023-jw}
I-Chun Chern, Steffi Chern, Shiqi Chen, Weizhe Yuan, Kehua Feng, Chunting Zhou,
  Junxian He, Graham Neubig, and Pengfei Liu.
\newblock {FacTool}: Factuality detection in generative {AI} -- a tool
  augmented framework for multi-task and multi-domain scenarios.
\newblock \emph{arXiv preprint arXiv:307.13528}, July 2023.

\bibitem[Chiang et~al.(2023)Chiang, Li, Lin, Sheng, Wu, Zhang, Zheng, Zhuang,
  Zhuang, Gonzalez, Stoica, and Xing]{Chiang2023-xe}
Wei-Lin Chiang, Zhuohan Li, Zi~Lin, Ying Sheng, Zhanghao Wu, Hao Zhang, Lianmin
  Zheng, Siyuan Zhuang, Yonghao Zhuang, Joseph~E. Gonzalez, Ion Stoica, and
  Eric~P. Xing.
\newblock Vicuna: An open-source chatbot impressing gpt-4 with 90\%* chatgpt
  quality.
\newblock \url{https://lmsys.org/blog/2023-03-30-vicuna/}, March 2023.

\bibitem[Chowdhery et~al.(2022)Chowdhery, Narang, Devlin, Bosma, Mishra,
  Roberts, Barham, Chung, Sutton, Gehrmann, Schuh, Shi, Tsvyashchenko, Maynez,
  Rao, Barnes, Tay, Shazeer, Prabhakaran, Reif, Du, Hutchinson, Pope, Bradbury,
  Austin, Isard, Gur-Ari, Yin, Duke, Levskaya, Ghemawat, Dev, Michalewski,
  Garcia, Misra, Robinson, Fedus, Zhou, Ippolito, Luan, Lim, Zoph, Spiridonov,
  Sepassi, Dohan, Agrawal, Omernick, Dai, Pillai, Pellat, Lewkowycz, Moreira,
  Child, Polozov, Lee, Zhou, Wang, Saeta, Diaz, Firat, Catasta, Wei,
  Meier-Hellstern, Eck, Dean, Petrov, and Fiedel]{Chowdhery2022-yw}
Aakanksha Chowdhery, Sharan Narang, Jacob Devlin, Maarten Bosma, Gaurav Mishra,
  Adam Roberts, Paul Barham, Hyung~Won Chung, Charles Sutton, Sebastian
  Gehrmann, Parker Schuh, Kensen Shi, Sasha Tsvyashchenko, Joshua Maynez,
  Abhishek Rao, Parker Barnes, Yi~Tay, Noam Shazeer, Vinodkumar Prabhakaran,
  Emily Reif, Nan Du, Ben Hutchinson, Reiner Pope, James Bradbury, Jacob
  Austin, Michael Isard, Guy Gur-Ari, Pengcheng Yin, Toju Duke, Anselm
  Levskaya, Sanjay Ghemawat, Sunipa Dev, Henryk Michalewski, Xavier Garcia,
  Vedant Misra, Kevin Robinson, Liam Fedus, Denny Zhou, Daphne Ippolito, David
  Luan, Hyeontaek Lim, Barret Zoph, Alexander Spiridonov, Ryan Sepassi, David
  Dohan, Shivani Agrawal, Mark Omernick, Andrew~M Dai,
  Thanumalayan~Sankaranarayana Pillai, Marie Pellat, Aitor Lewkowycz, Erica
  Moreira, Rewon Child, Oleksandr Polozov, Katherine Lee, Zongwei Zhou, Xuezhi
  Wang, Brennan Saeta, Mark Diaz, Orhan Firat, Michele Catasta, Jason Wei,
  Kathy Meier-Hellstern, Douglas Eck, Jeff Dean, Slav Petrov, and Noah Fiedel.
\newblock {PaLM}: Scaling language modeling with pathways.
\newblock \emph{arXiv preprint arXiv:2204.02311}, April 2022.

\bibitem[Clark \& Gardner(2018)Clark and Gardner]{Clark2018-oq}
Christopher Clark and Matt Gardner.
\newblock Simple and effective {Multi-Paragraph} reading comprehension.
\newblock In \emph{Proceedings of the 56th Annual Meeting of the Association
  for Computational Linguistics (Volume 1: Long Papers)}, pp.\  845--855,
  Melbourne, Australia, July 2018. Association for Computational Linguistics.

\bibitem[Clark et~al.(2020)Clark, Luong, Le, and Manning]{Clark2020-vh}
Kevin Clark, Minh-Thang Luong, Quoc~V Le, and Christopher~D Manning.
\newblock {ELECTRA}: Pre-training text encoders as discriminators rather than
  generators.
\newblock In \emph{International Conference on Learning Representations}, March
  2020.

\bibitem[Clark et~al.(2016)Clark, Etzioni, Khot, Sabharwal, Tafjord, Turney,
  and Khashabi]{Clark2016-xg}
Peter Clark, Oren Etzioni, Tushar Khot, Ashish Sabharwal, Oyvind Tafjord, Peter
  Turney, and Daniel Khashabi.
\newblock Combining retrieval, statistics, and inference to answer elementary
  science questions.
\newblock In \emph{{AAAI} Conference on Artificial Intelligence}, volume~30.
  Association for the Advancement of Artificial Intelligence, 2016.

\bibitem[Clark et~al.(2018)Clark, Cowhey, Etzioni, Khot, Sabharwal, Schoenick,
  and Tafjord]{Clark2018-gy}
Peter Clark, Isaac Cowhey, Oren Etzioni, Tushar Khot, Ashish Sabharwal, Carissa
  Schoenick, and Oyvind Tafjord.
\newblock Think you have solved question answering? try {ARC}, the {AI2}
  reasoning challenge.
\newblock \emph{arXiv preprint arXiv:1803.05457}, 2018.

\bibitem[Dem{\v s}ar(2006)]{Demsar2006-ze}
Janez Dem{\v s}ar.
\newblock Statistical comparisons of classifiers over multiple data sets.
\newblock \emph{Journal of Machine Learning Research}, 7:\penalty0 1--30, 2006.

\bibitem[Dettmers et~al.(2022)Dettmers, Lewis, Belkada, and
  Zettlemoyer]{Dettmers2022-te}
Tim Dettmers, Mike Lewis, Younes Belkada, and Luke Zettlemoyer.
\newblock {LLM.int8()}: 8-bit matrix multiplication for transformers at scale.
\newblock In \emph{36th Conference on Neural Information Processing Systems},
  pp.\  30318--30332, August 2022.

\bibitem[DeYoung et~al.(2020)DeYoung, Jain, Rajani, Lehman, Xiong, Socher, and
  Wallace]{DeYoung2020-cx}
Jay DeYoung, Sarthak Jain, Nazneen~Fatema Rajani, Eric Lehman, Caiming Xiong,
  Richard Socher, and Byron~C Wallace.
\newblock {ERASER}: A benchmark to evaluate rationalized {NLP} models.
\newblock In \emph{Proceedings of the 58th Annual Meeting of the Association
  for Computational Linguistics}, pp.\  4443--4458. Association for
  Computational Linguistics, 2020.

\bibitem[Dua et~al.(2019)Dua, Wang, Dasigi, Stanovsky, Singh, and
  Gardner]{Dua2019-td}
Dheeru Dua, Yizhong Wang, Pradeep Dasigi, Gabriel Stanovsky, Sameer Singh, and
  Matt Gardner.
\newblock {{DROP}}: A reading comprehension benchmark requiring discrete
  reasoning over paragraphs.
\newblock In \emph{Proceedings of the 2019 Conference of the North {A}merican
  Chapter of the Association for Computational Linguistics: Human Language
  Technologies, Volume 1 (Long and Short Papers)}, pp.\  2368--2378,
  Minneapolis, Minnesota, June 2019. Association for Computational Linguistics.

\bibitem[Ferguson et~al.(2020)Ferguson, Gardner, Hajishirzi, Khot, and
  Dasigi]{Ferguson2020-hv}
James Ferguson, Matt Gardner, Hannaneh Hajishirzi, Tushar Khot, and Pradeep
  Dasigi.
\newblock {IIRC}: A dataset of incomplete information reading comprehension
  questions.
\newblock In \emph{Proceedings of the 2020 Conference on Empirical Methods in
  Natural Language Processing ({EMNLP})}, pp.\  1137--1147, Stroudsburg, PA,
  USA, November 2020. Association for Computational Linguistics.

\bibitem[Fu et~al.(2023)Fu, Ng, Jiang, and Liu]{Fu2023-ir}
Jinlan Fu, See-Kiong Ng, Zhengbao Jiang, and Pengfei Liu.
\newblock {GPTScore}: Evaluate as you desire.
\newblock \emph{arXiv preprint arXiv:2302.04166}, February 2023.

\bibitem[Geva et~al.(2021)Geva, Khashabi, Segal, Khot, Roth, and
  Berant]{Geva2021-sl}
Mor Geva, Daniel Khashabi, Elad Segal, Tushar Khot, Dan Roth, and Jonathan
  Berant.
\newblock Did aristotle use a laptop? a question answering benchmark with
  implicit reasoning strategies.
\newblock \emph{Transactions of the Association for Computational Linguistics},
  9:\penalty0 346--361, 2021.

\bibitem[Hartill et~al.(2023)Hartill, TAN, Witbrock, and
  Riddle]{Hartill2023-pf}
Tim Hartill, Neset TAN, Michael Witbrock, and Patricia~J Riddle.
\newblock Teaching smaller language models to generalise to unseen
  compositional questions.
\newblock \emph{Transactions on Machine Learning Research}, August 2023.

\bibitem[Herbold(2020)]{Herbold2020-vn}
Steffen Herbold.
\newblock Autorank: A python package for automated ranking of classifiers.
\newblock \emph{Journal of Open Source Software}, 5\penalty0 (48):\penalty0
  2173, April 2020.

\bibitem[Holtzman et~al.(2019)Holtzman, Buys, Du, Forbes, and
  Choi]{Holtzman2019-nw}
Ari Holtzman, Jan Buys, Li~Du, Maxwell Forbes, and Yejin Choi.
\newblock The curious case of neural text degeneration.
\newblock In \emph{International Conference on Learning Representations},
  September 2019.

\bibitem[Hsieh et~al.(2023)Hsieh, Li, Yeh, Nakhost, Fujii, Ratner, Krishna,
  Lee, and Pfister]{Hsieh2023-my}
Cheng-Yu Hsieh, Chun-Liang Li, Chih-Kuan Yeh, Hootan Nakhost, Yasuhisa Fujii,
  Alexander Ratner, Ranjay Krishna, Chen-Yu Lee, and Tomas Pfister.
\newblock Distilling {Step-by-Step}! outperforming larger language models with
  less training data and smaller model sizes.
\newblock In \emph{Findings of the Association for Computational Linguistics:
  {ACL} 2023}, pp.\  8003--8017. Association for Computational Linguistics,
  2023.

\bibitem[Inoue et~al.(2020)Inoue, Stenetorp, and Inui]{Inoue2020-by}
Naoya Inoue, Pontus Stenetorp, and Kentaro Inui.
\newblock {R4C}: A benchmark for evaluating {RC} systems to get the right
  answer for the right reason.
\newblock In \emph{Proceedings of the 58th Annual Meeting of the Association
  for Computational Linguistics}, pp.\  6740--6750. Association for
  Computational Linguistics., 2020.

\bibitem[Izacard et~al.(2022)Izacard, Caron, Hosseini, Riedel, Bojanowski,
  Joulin, and Grave]{Izacard2022-hu}
Gautier Izacard, Mathilde Caron, Lucas Hosseini, Sebastian Riedel, Piotr
  Bojanowski, Armand Joulin, and Edouard Grave.
\newblock Unsupervised dense information retrieval with contrastive learning.
\newblock \emph{Transactions on Machine Learning Research}, August 2022.

\bibitem[Jhamtani \& Clark(2020)Jhamtani and Clark]{Jhamtani2020-dh}
Harsh Jhamtani and Peter Clark.
\newblock Learning to explain: Datasets and models for identifying valid
  reasoning chains in multihop {Question-Answering}.
\newblock In \emph{Proceedings of the 2020 Conference on Empirical Methods in
  Natural Language Processing}, pp.\  137--150, Online, 2020. Association for
  Computational Linguistics.

\bibitem[Jiang et~al.(2020)Jiang, Bordia, Zhong, Dognin, Singh, and
  Bansal]{Jiang2020-on}
Yichen Jiang, Shikha Bordia, Zheng Zhong, Charles Dognin, Maneesh Singh, and
  Mohit Bansal.
\newblock {HoVer}: A dataset for {Many-Hop} fact extraction and claim
  verification.
\newblock In \emph{Findings of the Association for Computational Linguistics:
  {EMNLP} 2020}, pp.\  3441--3460. Association for Computational Linguistics,
  2020.

\bibitem[Karpukhin et~al.(2020)Karpukhin, Oguz, Min, Lewis, Wu, Edunov, Chen,
  and Yih]{Karpukhin2020-wa}
Vladimir Karpukhin, Barlas Oguz, Sewon Min, Patrick Lewis, Ledell Wu, Sergey
  Edunov, Danqi Chen, and Wen-Tau Yih.
\newblock Dense passage retrieval for {Open-Domain} question answering.
\newblock In \emph{Proceedings of the 2020 Conference on Empirical Methods in
  Natural Language Processing ({EMNLP})}, pp.\  6769--6781, Online, November
  2020. Association for Computational Linguistics.

\bibitem[Khashabi et~al.(2020)Khashabi, Min, Khot, Sabharwal, Tafjord, Clark,
  and Hajishirzi]{Khashabi2020-gq}
Daniel Khashabi, Sewon Min, Tushar Khot, Ashish Sabharwal, Oyvind Tafjord,
  Peter Clark, and Hannaneh Hajishirzi.
\newblock {{UNIFIEDQA}}: Crossing format boundaries with a single {QA} system.
\newblock In \emph{Findings of the Association for Computational Linguistics:
  {EMNLP} 2020}, pp.\  1896--1907, Online, 2020. Association for Computational
  Linguistics.

\bibitem[Khattab et~al.(2021)Khattab, Potts, and Zaharia]{Khattab2021-jf}
Omar Khattab, Christopher Potts, and Matei Zaharia.
\newblock Baleen: Robust multi-hop reasoning at scale via condensed retrieval.
\newblock In \emph{Advances in Neural Information Processing Systems, 34}, pp.\
   27670--27682, 2021.

\bibitem[Khot et~al.(2020)Khot, Clark, Guerquin, Jansen, and
  Sabharwal]{Khot2020-sv}
Tushar Khot, Peter Clark, Michal Guerquin, Peter Jansen, and Ashish Sabharwal.
\newblock {QASC}: A dataset for question answering via sentence composition.
\newblock In \emph{Proceedings of the {AAAI} Conference on Artificial
  Intelligence}, volume 34(05), pp.\  8082--8090. Association for the
  Advancement of Artificial Intelligence, 2020.

\bibitem[K{\"o}pf et~al.(2023)K{\"o}pf, Kilcher, von R{\"u}tte, Anagnostidis,
  Tam, Stevens, Barhoum, Duc, Stanley, Nagyfi, Shahul, Suri, Glushkov,
  Dantuluri, Maguire, Schuhmann, Nguyen, and Mattick]{Kopf2023-bx}
Andreas K{\"o}pf, Yannic Kilcher, Dimitri von R{\"u}tte, Sotiris Anagnostidis,
  Zhi-Rui Tam, Keith Stevens, Abdullah Barhoum, Nguyen~Minh Duc, Oliver
  Stanley, Rich{\'a}rd Nagyfi, E~S Shahul, Sameer Suri, David Glushkov, Arnav
  Dantuluri, Andrew Maguire, Christoph Schuhmann, Huu Nguyen, and Alexander
  Mattick.
\newblock {OpenAssistant} conversations -- democratizing large language model
  alignment.
\newblock \emph{arXiv preprint arXiv:2304.07327}, April 2023.

\bibitem[Lewis et~al.(2020)Lewis, Liu, Goyal, Ghazvininejad, Mohamed, Levy,
  Stoyanov, and Zettlemoyer]{Lewis2020-gt}
Mike Lewis, Yinhan Liu, Naman Goyal, Marjan Ghazvininejad, Abdelrahman Mohamed,
  Omer Levy, Veselin Stoyanov, and Luke Zettlemoyer.
\newblock {{BART}}: Denoising {Sequence-to-Sequence} pre-training for natural
  language generation, translation, and comprehension.
\newblock In \emph{Proceedings of the 58th Annual Meeting of the Association
  for Computational Linguistics}, pp.\  7871--7880, Online, 2020. Association
  for Computational Linguistics.

\bibitem[Lewis et~al.(2021)Lewis, Stenetorp, and Riedel]{Lewis2021-ia}
Patrick Lewis, Pontus Stenetorp, and Sebastian Riedel.
\newblock Question and answer {Test-Train} overlap in {Open-Domain} question
  answering datasets.
\newblock In \emph{Proceedings of the 16th Conference of the European Chapter
  of the Association for Computational Linguistics: Main Volume}, pp.\
  1000--1008, Online, 2021. Association for Computational Linguistics.

\bibitem[Li et~al.(2023)Li, Hessel, Yu, Ren, Chang, and Choi]{Li2023-ss}
Liunian~Harold Li, Jack Hessel, Youngjae Yu, Xiang Ren, Kai-Wei Chang, and
  Yejin Choi.
\newblock Symbolic {Chain-of-Thought} distillation: Small models can also
  ``think'' step-by-step.
\newblock In \emph{Proceedings of the 61st Annual Meeting of the Association
  for Computational Linguistics (Volume 1: Long Papers)}, pp.\  2665--2679.
  Association for Computational Linguistics, June 2023.

\bibitem[Lin et~al.(2022{\natexlab{a}})Lin, Nogueira, and Yates]{Lin2022-jq}
Jimmy Lin, Rodrigo Nogueira, and Andrew Yates.
\newblock \emph{Pretrained Transformers for Text Ranking: {BERT} and Beyond}.
\newblock Springer Nature, June 2022{\natexlab{a}}.

\bibitem[Lin et~al.(2022{\natexlab{b}})Lin, Hilton, and Evans]{Lin2022-bq}
Stephanie Lin, Jacob Hilton, and Owain Evans.
\newblock {TruthfulQA}: Measuring how models mimic human falsehoods.
\newblock In \emph{Proceedings of the 60th Annual Meeting of the Association
  for Computational Linguistics (Volume 1: Long Papers)}, pp.\  3214--3252,
  Stroudsburg, PA, USA, May 2022{\natexlab{b}}. Association for Computational
  Linguistics.

\bibitem[Liu et~al.(2022)Liu, Liu, Lu, Welleck, West, Le~Bras, Choi, and
  Hajishirzi]{Liu2022-bc}
Jiacheng Liu, Alisa Liu, Ximing Lu, Sean Welleck, Peter West, Ronan Le~Bras,
  Yejin Choi, and Hannaneh Hajishirzi.
\newblock Generated knowledge prompting for commonsense reasoning.
\newblock In \emph{Proceedings of the 60th Annual Meeting of the Association
  for Computational Linguistics (Volume 1: Long Papers)}, pp.\  3154--3169,
  Stroudsburg, PA, USA, May 2022. Association for Computational Linguistics.

\bibitem[Lourie et~al.(2021)Lourie, Le~Bras, Bhagavatula, and
  Choi]{Lourie2021-kb}
Nicholas Lourie, Ronan Le~Bras, Chandra Bhagavatula, and Yejin Choi.
\newblock {UNICORN} on {RAINBOW}: A universal commonsense reasoning model on a
  new multitask benchmark.
\newblock In \emph{Proceedings of the {AAAI} Conference on Artificial
  Intelligence}, volume~35 of \emph{15}, pp.\  13480--13488, May 2021.

\bibitem[Madaan et~al.(2023)Madaan, Tandon, Gupta, Hallinan, Gao, Wiegreffe,
  Alon, Dziri, Prabhumoye, Yang, Welleck, Majumder, Gupta, Yazdanbakhsh, and
  Clark]{Madaan2023-rm}
Aman Madaan, Niket Tandon, Prakhar Gupta, Skyler Hallinan, Luyu Gao, Sarah
  Wiegreffe, Uri Alon, Nouha Dziri, Shrimai Prabhumoye, Yiming Yang, Sean
  Welleck, Bodhisattwa~Prasad Majumder, Shashank Gupta, Amir Yazdanbakhsh, and
  Peter Clark.
\newblock Self-refine: Iterative refinement with self-feedback.
\newblock \emph{arXiv preprint arXiv:2303.17651}, March 2023.

\bibitem[Magister et~al.(2023)Magister, Mallinson, Adamek, Malmi, and
  Severyn]{Magister2023-uf}
Lucie~Charlotte Magister, Jonathan Mallinson, Jakub Adamek, Eric Malmi, and
  Aliaksei Severyn.
\newblock Teaching small language models to reason.
\newblock In \emph{Proceedings of the 61st Annual Meeting of the Association
  for Computational Linguistics (Volume 2: Short Papers)}, pp.\  1773--1781,
  Toronto, Canada, July 2023. Association for Computational Linguistics.

\bibitem[Manakul et~al.(2023)Manakul, Liusie, and Gales]{Manakul2023-db}
Potsawee Manakul, Adian Liusie, and Mark J~F Gales.
\newblock {SelfCheckGPT}: Zero-resource black-box hallucination detection for
  generative large language models.
\newblock \emph{arXiv preprint arXiv:2303.08896}, March 2023.

\bibitem[Min et~al.(2023)Min, Krishna, Lyu, Lewis, Yih, Koh, Iyyer,
  Zettlemoyer, and Hajishirzi]{Min2023-fc}
Sewon Min, Kalpesh Krishna, Xinxi Lyu, Mike Lewis, Wen-Tau Yih, Pang~Wei Koh,
  Mohit Iyyer, Luke Zettlemoyer, and Hannaneh Hajishirzi.
\newblock {FActScore}: Fine-grained atomic evaluation of factual precision in
  long form text generation.
\newblock \emph{arXiv preprint arXiv:2305.14251}, May 2023.

\bibitem[Mitra \& Craswell(2018)Mitra and Craswell]{Mitra2018-mr}
Bhaskar Mitra and Nick Craswell.
\newblock An introduction to neural information retrieval.
\newblock \emph{Foundations and Trends\textregistered{} in Information
  Retrieval}, 13\penalty0 (1):\penalty0 1--126, 2018.

\bibitem[Moravec(1988)]{Moravec1988-yw}
Hans Moravec.
\newblock \emph{Mind Children: The Future of Robot and Human Intelligence}.
\newblock Harvard University Press, 1988.

\bibitem[Onoe et~al.(2021)Onoe, Zhang, Choi, and Durrett]{Onoe2021-il}
Yasumasa Onoe, Michael J~Q Zhang, Eunsol Choi, and Greg Durrett.
\newblock {CREAK}: A dataset for commonsense reasoning over entity knowledge.
\newblock In \emph{Thirty-fifth Conference on Neural Information Processing
  Systems Datasets and Benchmarks Track (Round 2)}, November 2021.

\bibitem[{OpenAI}(2023)]{OpenAI2023-cv}
{OpenAI}.
\newblock {GPT-4} technical report.
\newblock \emph{arXiv preprint arXiv:2303.08774}, March 2023.

\bibitem[Ouyang et~al.(2022)Ouyang, Wu, Jiang, Almeida, Wainwright, Mishkin,
  Zhang, Agarwal, Slama, Ray, Schulman, Hilton, Kelton, Miller, Simens, Askell,
  Welinder, Christiano, Leike, and Lowe]{Ouyang2022-ti}
Long Ouyang, Jeff Wu, Xu~Jiang, Diogo Almeida, Carroll~L Wainwright, Pamela
  Mishkin, Chong Zhang, Sandhini Agarwal, Katarina Slama, Alex Ray, John
  Schulman, Jacob Hilton, Fraser Kelton, Luke Miller, Maddie Simens, Amanda
  Askell, Peter Welinder, Paul Christiano, Jan Leike, and Ryan Lowe.
\newblock Training language models to follow instructions with human feedback.
\newblock In \emph{Advances in Neural Information Processing Systems}, 35, pp.\
   27730--27744, 2022.

\bibitem[Pan et~al.(2023)Pan, Yao, Zhang, Yu, Yu, and Chen]{Pan2023-sz}
Xiaoman Pan, Wenlin Yao, Hongming Zhang, Dian Yu, Dong Yu, and Jianshu Chen.
\newblock {Knowledge-in-Context}: Towards knowledgeable {Semi-Parametric}
  language models.
\newblock In \emph{The Eleventh International Conference on Learning
  Representations}, 2023.

\bibitem[Piktus et~al.(2021)Piktus, Petroni, Karpukhin, Okhonko, Broscheit,
  Izacard, Lewis, O{\u g}uz, Grave, Yih, and Riedel]{Piktus2021-lu}
Aleksandra Piktus, Fabio Petroni, Vladimir Karpukhin, Dmytro Okhonko, Samuel
  Broscheit, Gautier Izacard, Patrick Lewis, Barlas O{\u g}uz, Edouard Grave,
  Wen-Tau Yih, and Sebastian Riedel.
\newblock The web is your oyster - knowledge-intensive {NLP} against a very
  large web corpus.
\newblock \emph{arXiv preprint arXiv:2112.09924}, December 2021.

\bibitem[Radford et~al.(2018)Radford, Narasimhan, Salimans, and
  Sutskever]{Radford2018-kr}
Alec Radford, Karthik Narasimhan, Tim Salimans, and Ilya Sutskever.
\newblock Improving language understanding by generative pre-training.
\newblock
  \url{http://openai-assets.s3.amazonaws.com/research-covers/language-unsupervised/language_understanding_paper.pdf},
  2018.

\bibitem[Sap et~al.(2019)Sap, Le~Bras, Allaway, Bhagavatula, Lourie, Rashkin,
  Roof, Smith, and Choi]{Sap2019-tk}
Maarten Sap, Ronan Le~Bras, Emily Allaway, Chandra Bhagavatula, Nicholas
  Lourie, Hannah Rashkin, Brendan Roof, Noah~A Smith, and Yejin Choi.
\newblock {ATOMIC}: An atlas of machine commonsense for if-then reasoning.
\newblock In \emph{Proceedings of the {AAAI} Conference on Artificial
  Intelligence}, 33(01), pp.\  3027--3035, 2019.

\bibitem[Shridhar et~al.(2023)Shridhar, Stolfo, and Sachan]{Shridhar2023-qb}
Kumar Shridhar, Alessandro Stolfo, and Mrinmaya Sachan.
\newblock Distilling reasoning capabilities into smaller language models.
\newblock In \emph{Findings of the Association for Computational Linguistics:
  {ACL} 2023}, pp.\  7059--7073, Toronto, Canada, July 2023. Association for
  Computational Linguistics.

\bibitem[Shwartz et~al.(2020)Shwartz, West, Le~Bras, Bhagavatula, and
  Choi]{Shwartz2020-sn}
Vered Shwartz, Peter West, Ronan Le~Bras, Chandra Bhagavatula, and Yejin Choi.
\newblock Unsupervised commonsense question answering with self-talk.
\newblock In \emph{Proceedings of the 2020 Conference on Empirical Methods in
  Natural Language Processing ({EMNLP})}, pp.\  4615--4629, Stroudsburg, PA,
  USA, November 2020. Association for Computational Linguistics.

\bibitem[Si et~al.(2023)Si, Shi, Zhao, Zettlemoyer, and Boyd-Graber]{Si2023-kz}
Chenglei Si, Weijia Shi, Chen Zhao, Luke Zettlemoyer, and Jordan Boyd-Graber.
\newblock Mixture of prompt experts for generalizable and interpretable
  question answering.
\newblock \emph{arXiv preprint arXiv 2305.14628}, May 2023.

\bibitem[Sp{\"a}rck~Jones(1972)]{Sparck_Jones1972-dm}
Karen Sp{\"a}rck~Jones.
\newblock A statistical interpretation of term specificity and its application
  in retrieval.
\newblock \emph{Journal of Documentation}, 28\penalty0 (1):\penalty0 11--21,
  1972.

\bibitem[Speer et~al.(2017)Speer, Chin, and Havasi]{Speer2017-ll}
Robyn Speer, Joshua Chin, and Catherine Havasi.
\newblock {ConceptNet} 5.5: An open multilingual graph of general knowledge.
\newblock In \emph{Proceedings of the {AAAI} Conference on Artificial
  Intelligence}, 31(1), pp.\  4444--4451, 2017.

\bibitem[Srivastava et~al.(2022)Srivastava, Rastogi, Rao, Shoeb, Abid, Fisch,
  Brown, Santoro, Gupta, Garriga-Alonso, Kluska, Lewkowycz, Agarwal, Power,
  Ray, Warstadt, Kocurek, Safaya, Tazarv, Xiang, Parrish, Nie, Hussain, Askell,
  Dsouza, Slone, Rahane, Iyer, Andreassen, Madotto, Santilli, Stuhlm{\"u}ller,
  Dai, La, Lampinen, Zou, Jiang, Chen, Vuong, Gupta, Gottardi, Norelli,
  Venkatesh, Gholamidavoodi, Tabassum, Menezes, Kirubarajan, Mullokandov,
  Sabharwal, Herrick, Efrat, Erdem, Karaka{\c s}, Ryan~Roberts, Loe, Zoph,
  Bojanowski, {\"O}zyurt, Hedayatnia, Neyshabur, Inden, Stein, Ekmekci, Lin,
  Howald, Diao, Dour, Stinson, Argueta, Ram{\'\i}rez, Singh, Rathkopf, Meng,
  Baral, Wu, Callison-Burch, Waites, Voigt, Manning, Potts, Ramirez, Rivera,
  Siro, Raffel, Ashcraft, Garbacea, Sileo, Garrette, Hendrycks, Kilman, Roth,
  Freeman, Khashabi, Levy, Gonz{\'a}lez, Perszyk, Hernandez, Chen, Ippolito,
  Gilboa, Dohan, Drakard, Jurgens, Datta, Ganguli, Emelin, Kleyko, Yuret, Chen,
  Tam, Hupkes, Misra, Buzan, Mollo, Yang, Lee, Shutova, Cubuk, Segal, Hagerman,
  Barnes, Donoway, Pavlick, Rodola, Lam, Chu, Tang, Erdem, Chang, Chi, Dyer,
  Jerzak, Kim, Manyasi, Zheltonozhskii, Xia, Siar, Mart{\'\i}nez-Plumed,
  Happ{\'e}, Chollet, Rong, Mishra, Winata, de~Melo, Kruszewski, Parascandolo,
  Mariani, Wang, Jaimovitch-L{\'o}pez, Betz, Gur-Ari, Galijasevic, Kim,
  Rashkin, Hajishirzi, Mehta, Bogar, Shevlin, Sch{\"u}tze, Yakura, Zhang, Wong,
  Ng, Noble, Jumelet, Geissinger, Kernion, Hilton, Lee, Fisac, Simon, Koppel,
  Zheng, Zou, Koco{\'n}, Thompson, Kaplan, Radom, Sohl-Dickstein, Phang, Wei,
  Yosinski, Novikova, Bosscher, Marsh, Kim, Taal, Engel, Alabi, Xu, Song, Tang,
  Waweru, Burden, Miller, Balis, Berant, Frohberg, Rozen, Hernandez-Orallo,
  Boudeman, Jones, Tenenbaum, Rule, Chua, Kanclerz, Livescu, Krauth,
  Gopalakrishnan, Ignatyeva, Markert, Dhole, Gimpel, Omondi, Mathewson,
  Chiafullo, Shkaruta, Shridhar, McDonell, Richardson, Reynolds, Gao, Zhang,
  Dugan, Qin, Contreras-Ochando, Morency, Moschella, Lam, Noble, Schmidt, He,
  Col{\'o}n, Metz, {\c S}enel, Bosma, Sap, ter Hoeve, Farooqi, Faruqui,
  Mazeika, Baturan, Marelli, Maru, Quintana, Tolkiehn, Giulianelli, Lewis,
  Potthast, Leavitt, Hagen, Schubert, Baitemirova, Arnaud, McElrath, Yee,
  Cohen, Gu, Ivanitskiy, Starritt, Strube, Sw{\k e}drowski, Bevilacqua,
  Yasunaga, Kale, Cain, Xu, Suzgun, Tiwari, Bansal, Aminnaseri, Geva, Gheini,
  Mukund, Peng, Chi, Lee, Krakover, Cameron, Roberts, Doiron, Nangia, Deckers,
  Muennighoff, Keskar, Iyer, Constant, Fiedel, Wen, Zhang, Agha, Elbaghdadi,
  Levy, Evans, Casares, Doshi, Fung, Liang, Vicol, Alipoormolabashi, Liao,
  Liang, Chang, Eckersley, Htut, Hwang, Mi{\l}kowski, Patil, Pezeshkpour, Oli,
  Mei, Lyu, Chen, Banjade, Rudolph, Gabriel, Habacker, Delgado, Milli{\`e}re,
  Garg, Barnes, Saurous, Arakawa, Raymaekers, Frank, Sikand, Novak, Sitelew,
  LeBras, Liu, Jacobs, Zhang, Salakhutdinov, Chi, Lee, Stovall, Teehan, Yang,
  Singh, Mohammad, Anand, Dillavou, Shleifer, Wiseman, Gruetter, Bowman,
  Schoenholz, Han, Kwatra, Rous, Ghazarian, Ghosh, Casey, Bischoff, Gehrmann,
  Schuster, Sadeghi, Hamdan, Zhou, Srivastava, Shi, Singh, Asaadi, Gu,
  Pachchigar, Toshniwal, Upadhyay, {Shyamolima}, {Debnath}, Shakeri, Thormeyer,
  Melzi, Reddy, Makini, Lee, Torene, Hatwar, Dehaene, Divic, Ermon, Biderman,
  Lin, Prasad, Piantadosi, Shieber, Misherghi, Kiritchenko, Mishra, Linzen,
  Schuster, Li, Yu, Ali, Hashimoto, Wu, Desbordes, Rothschild, Phan, Wang,
  Nkinyili, Schick, Kornev, Telleen-Lawton, Tunduny, Gerstenberg, Chang,
  Neeraj, Khot, Shultz, Shaham, Misra, Demberg, Nyamai, Raunak, Ramasesh,
  Prabhu, Padmakumar, Srikumar, Fedus, Saunders, Zhang, Vossen, Ren, Tong,
  Zhao, Wu, Shen, Yaghoobzadeh, Lakretz, Song, Bahri, Choi, Yang, Hao, Chen,
  Belinkov, Hou, Hou, Bai, Seid, Zhao, Wang, Wang, Wang, and
  Wu]{Srivastava2022-kp}
Aarohi Srivastava, Abhinav Rastogi, Abhishek Rao, Abu Awal~Md Shoeb, Abubakar
  Abid, Adam Fisch, Adam~R Brown, Adam Santoro, Aditya Gupta, Adri{\`a}
  Garriga-Alonso, Agnieszka Kluska, Aitor Lewkowycz, Akshat Agarwal, Alethea
  Power, Alex Ray, Alex Warstadt, Alexander~W Kocurek, Ali Safaya, Ali Tazarv,
  Alice Xiang, Alicia Parrish, Allen Nie, Aman Hussain, Amanda Askell, Amanda
  Dsouza, Ambrose Slone, Ameet Rahane, Anantharaman~S Iyer, Anders Andreassen,
  Andrea Madotto, Andrea Santilli, Andreas Stuhlm{\"u}ller, Andrew Dai, Andrew
  La, Andrew Lampinen, Andy Zou, Angela Jiang, Angelica Chen, Anh Vuong,
  Animesh Gupta, Anna Gottardi, Antonio Norelli, Anu Venkatesh, Arash
  Gholamidavoodi, Arfa Tabassum, Arul Menezes, Arun Kirubarajan, Asher
  Mullokandov, Ashish Sabharwal, Austin Herrick, Avia Efrat, Aykut Erdem, Ayla
  Karaka{\c s}, B~Ryan~Roberts, Bao~Sheng Loe, Barret Zoph, Bart{\l}omiej
  Bojanowski, Batuhan {\"O}zyurt, Behnam Hedayatnia, Behnam Neyshabur, Benjamin
  Inden, Benno Stein, Berk Ekmekci, Bill~Yuchen Lin, Blake Howald, Cameron
  Diao, Cameron Dour, Catherine Stinson, Cedrick Argueta, C{\'e}sar~Ferri
  Ram{\'\i}rez, Chandan Singh, Charles Rathkopf, Chenlin Meng, Chitta Baral,
  Chiyu Wu, Chris Callison-Burch, Chris Waites, Christian Voigt, Christopher~D
  Manning, Christopher Potts, Cindy Ramirez, Clara~E Rivera, Clemencia Siro,
  Colin Raffel, Courtney Ashcraft, Cristina Garbacea, Damien Sileo, Dan
  Garrette, Dan Hendrycks, Dan Kilman, Dan Roth, Daniel Freeman, Daniel
  Khashabi, Daniel Levy, Daniel~Mosegu{\'\i} Gonz{\'a}lez, Danielle Perszyk,
  Danny Hernandez, Danqi Chen, Daphne Ippolito, Dar Gilboa, David Dohan, David
  Drakard, David Jurgens, Debajyoti Datta, Deep Ganguli, Denis Emelin, Denis
  Kleyko, Deniz Yuret, Derek Chen, Derek Tam, Dieuwke Hupkes, Diganta Misra,
  Dilyar Buzan, Dimitri~Coelho Mollo, Diyi Yang, Dong-Ho Lee, Ekaterina
  Shutova, Ekin~Dogus Cubuk, Elad Segal, Eleanor Hagerman, Elizabeth Barnes,
  Elizabeth Donoway, Ellie Pavlick, Emanuele Rodola, Emma Lam, Eric Chu, Eric
  Tang, Erkut Erdem, Ernie Chang, Ethan~A Chi, Ethan Dyer, Ethan Jerzak, Ethan
  Kim, Eunice~Engefu Manyasi, Evgenii Zheltonozhskii, Fanyue Xia, Fatemeh Siar,
  Fernando Mart{\'\i}nez-Plumed, Francesca Happ{\'e}, Francois Chollet, Frieda
  Rong, Gaurav Mishra, Genta~Indra Winata, Gerard de~Melo, Germ{\'a}n
  Kruszewski, Giambattista Parascandolo, Giorgio Mariani, Gloria Wang, Gonzalo
  Jaimovitch-L{\'o}pez, Gregor Betz, Guy Gur-Ari, Hana Galijasevic, Hannah Kim,
  Hannah Rashkin, Hannaneh Hajishirzi, Harsh Mehta, Hayden Bogar, Henry
  Shevlin, Hinrich Sch{\"u}tze, Hiromu Yakura, Hongming Zhang, Hugh~Mee Wong,
  Ian Ng, Isaac Noble, Jaap Jumelet, Jack Geissinger, Jackson Kernion, Jacob
  Hilton, Jaehoon Lee, Jaime~Fern{\'a}ndez Fisac, James~B Simon, James Koppel,
  James Zheng, James Zou, Jan Koco{\'n}, Jana Thompson, Jared Kaplan, Jarema
  Radom, Jascha Sohl-Dickstein, Jason Phang, Jason Wei, Jason Yosinski,
  Jekaterina Novikova, Jelle Bosscher, Jennifer Marsh, Jeremy Kim, Jeroen Taal,
  Jesse Engel, Jesujoba Alabi, Jiacheng Xu, Jiaming Song, Jillian Tang, Joan
  Waweru, John Burden, John Miller, John~U Balis, Jonathan Berant, J{\"o}rg
  Frohberg, Jos Rozen, Jose Hernandez-Orallo, Joseph Boudeman, Joseph Jones,
  Joshua~B Tenenbaum, Joshua~S Rule, Joyce Chua, Kamil Kanclerz, Karen Livescu,
  Karl Krauth, Karthik Gopalakrishnan, Katerina Ignatyeva, Katja Markert,
  Kaustubh~D Dhole, Kevin Gimpel, Kevin Omondi, Kory Mathewson, Kristen
  Chiafullo, Ksenia Shkaruta, Kumar Shridhar, Kyle McDonell, Kyle Richardson,
  Laria Reynolds, Leo Gao, Li~Zhang, Liam Dugan, Lianhui Qin, Lidia
  Contreras-Ochando, Louis-Philippe Morency, Luca Moschella, Lucas Lam, Lucy
  Noble, Ludwig Schmidt, Luheng He, Luis~Oliveros Col{\'o}n, Luke Metz,
  L{\"u}tfi~Kerem {\c S}enel, Maarten Bosma, Maarten Sap, Maartje ter Hoeve,
  Maheen Farooqi, Manaal Faruqui, Mantas Mazeika, Marco Baturan, Marco Marelli,
  Marco Maru, Maria Jose~Ram{\'\i}rez Quintana, Marie Tolkiehn, Mario
  Giulianelli, Martha Lewis, Martin Potthast, Matthew~L Leavitt, Matthias
  Hagen, M{\'a}ty{\'a}s Schubert, Medina~Orduna Baitemirova, Melody Arnaud,
  Melvin McElrath, Michael~A Yee, Michael Cohen, Michael Gu, Michael
  Ivanitskiy, Michael Starritt, Michael Strube, Micha{\l} Sw{\k e}drowski,
  Michele Bevilacqua, Michihiro Yasunaga, Mihir Kale, Mike Cain, Mimee Xu,
  Mirac Suzgun, Mo~Tiwari, Mohit Bansal, Moin Aminnaseri, Mor Geva, Mozhdeh
  Gheini, Varma~T Mukund, Nanyun Peng, Nathan Chi, Nayeon Lee, Neta Gur-Ari
  Krakover, Nicholas Cameron, Nicholas Roberts, Nick Doiron, Nikita Nangia,
  Niklas Deckers, Niklas Muennighoff, Nitish~Shirish Keskar, Niveditha~S Iyer,
  Noah Constant, Noah Fiedel, Nuan Wen, Oliver Zhang, Omar Agha, Omar
  Elbaghdadi, Omer Levy, Owain Evans, Pablo Antonio~Moreno Casares, Parth
  Doshi, Pascale Fung, Paul~Pu Liang, Paul Vicol, Pegah Alipoormolabashi,
  Peiyuan Liao, Percy Liang, Peter Chang, Peter Eckersley, Phu~Mon Htut, Pinyu
  Hwang, Piotr Mi{\l}kowski, Piyush Patil, Pouya Pezeshkpour, Priti Oli,
  Qiaozhu Mei, Qing Lyu, Qinlang Chen, Rabin Banjade, Rachel~Etta Rudolph,
  Raefer Gabriel, Rahel Habacker, Ram{\'o}n~Risco Delgado, Rapha{\"e}l
  Milli{\`e}re, Rhythm Garg, Richard Barnes, Rif~A Saurous, Riku Arakawa, Robbe
  Raymaekers, Robert Frank, Rohan Sikand, Roman Novak, Roman Sitelew, Ronan
  LeBras, Rosanne Liu, Rowan Jacobs, Rui Zhang, Ruslan Salakhutdinov, Ryan Chi,
  Ryan Lee, Ryan Stovall, Ryan Teehan, Rylan Yang, Sahib Singh, Saif~M
  Mohammad, Sajant Anand, Sam Dillavou, Sam Shleifer, Sam Wiseman, Samuel
  Gruetter, Samuel~R Bowman, Samuel~S Schoenholz, Sanghyun Han, Sanjeev Kwatra,
  Sarah~A Rous, Sarik Ghazarian, Sayan Ghosh, Sean Casey, Sebastian Bischoff,
  Sebastian Gehrmann, Sebastian Schuster, Sepideh Sadeghi, Shadi Hamdan, Sharon
  Zhou, Shashank Srivastava, Sherry Shi, Shikhar Singh, Shima Asaadi,
  Shixiang~Shane Gu, Shubh Pachchigar, Shubham Toshniwal, Shyam Upadhyay,
  {Shyamolima}, {Debnath}, Siamak Shakeri, Simon Thormeyer, Simone Melzi, Siva
  Reddy, Sneha~Priscilla Makini, Soo-Hwan Lee, Spencer Torene, Sriharsha
  Hatwar, Stanislas Dehaene, Stefan Divic, Stefano Ermon, Stella Biderman,
  Stephanie Lin, Stephen Prasad, Steven~T Piantadosi, Stuart~M Shieber, Summer
  Misherghi, Svetlana Kiritchenko, Swaroop Mishra, Tal Linzen, Tal Schuster,
  Tao Li, Tao Yu, Tariq Ali, Tatsu Hashimoto, Te-Lin Wu, Th{\'e}o Desbordes,
  Theodore Rothschild, Thomas Phan, Tianle Wang, Tiberius Nkinyili, Timo
  Schick, Timofei Kornev, Timothy Telleen-Lawton, Titus Tunduny, Tobias
  Gerstenberg, Trenton Chang, Trishala Neeraj, Tushar Khot, Tyler Shultz, Uri
  Shaham, Vedant Misra, Vera Demberg, Victoria Nyamai, Vikas Raunak, Vinay
  Ramasesh, Vinay~Uday Prabhu, Vishakh Padmakumar, Vivek Srikumar, William
  Fedus, William Saunders, William Zhang, Wout Vossen, Xiang Ren, Xiaoyu Tong,
  Xinran Zhao, Xinyi Wu, Xudong Shen, Yadollah Yaghoobzadeh, Yair Lakretz,
  Yangqiu Song, Yasaman Bahri, Yejin Choi, Yichi Yang, Yiding Hao, Yifu Chen,
  Yonatan Belinkov, Yu~Hou, Yufang Hou, Yuntao Bai, Zachary Seid, Zhuoye Zhao,
  Zijian Wang, Zijie~J Wang, Zirui Wang, and Ziyi Wu.
\newblock Beyond the imitation game: Quantifying and extrapolating the
  capabilities of language models.
\newblock \emph{arXiv preprint arXiv:2206.04615}, June 2022.

\bibitem[{Stability-AI}(2023)]{Stability-AI2023-mp}
{Stability-AI}.
\newblock {Stability AI releases StableVicuna, the AI World’s First Open
  Source RLHF LLM Chatbot}.
\newblock
  \url{https://stability.ai/blog/stablevicuna-open-source-rlhf-chatbot/}, April
  2023.
\newblock Accessed: 2023-7-5.

\bibitem[Talmor et~al.(2019)Talmor, Herzig, Lourie, and Berant]{Talmor2019-rm}
Alon Talmor, Jonathan Herzig, Nicholas Lourie, and Jonathan Berant.
\newblock {CommonsenseQA}: A question answering challenge targeting commonsense
  knowledge.
\newblock In \emph{Proceedings of the 2019 Conference of the North American
  Chapter of the Association for Computational Linguistics: Human Language
  Technologies, Volume 1 (Long and Short Papers)}, pp.\  4149--4158.
  Association for Computational Linguistics, 2019.

\bibitem[Taori et~al.(2023)Taori, Gulrajani, Zhang, Dubois, Li, Guestrin,
  Liang, and Hashimoto]{Taori2023-au}
Rohan Taori, Ishaan Gulrajani, Tianyi Zhang, Yann Dubois, Xuechen Li, Carlos
  Guestrin, Percy Liang, and Tatsunori~B Hashimoto.
\newblock Stanford alpaca: An instruction-following {LLaMA} model.
\newblock \url{https://github.com/tatsu-lab/stanford_alpaca}, 2023.

\bibitem[Thorne et~al.(2018)Thorne, Vlachos, Christodoulopoulos, and
  Mittal]{Thorne2018-ws}
James Thorne, Andreas Vlachos, Christos Christodoulopoulos, and Arpit Mittal.
\newblock {FEVER}: A large-scale dataset for fact extraction and
  {VERification}.
\newblock In \emph{Proceedings of the 2018 Conference of the North American
  Chapter of the Association for Computational Linguistics: Human Language
  Technologies, Volume 1 (Long Papers)}, Stroudsburg, PA, USA, 2018.
  Association for Computational Linguistics.

\bibitem[Touvron et~al.(2023)Touvron, Lavril, Izacard, Martinet, Lachaux,
  Lacroix, Rozi{\`e}re, Goyal, Hambro, Azhar, Rodriguez, Joulin, Grave, and
  Lample]{Touvron2023-nj}
Hugo Touvron, Thibaut Lavril, Gautier Izacard, Xavier Martinet, Marie-Anne
  Lachaux, Timoth{\'e}e Lacroix, Baptiste Rozi{\`e}re, Naman Goyal, Eric
  Hambro, Faisal Azhar, Aurelien Rodriguez, Armand Joulin, Edouard Grave, and
  Guillaume Lample.
\newblock {LLaMA}: Open and efficient foundation language models.
\newblock \emph{arXiv preprint arXiv:2302.13971}, February 2023.

\bibitem[Trivedi et~al.(2022)Trivedi, Balasubramanian, Khot, and
  Sabharwal]{Trivedi2022-mv}
Harsh Trivedi, Niranjan Balasubramanian, Tushar Khot, and Ashish Sabharwal.
\newblock {MuSiQue}: Multihop questions via single-hop question composition.
\newblock \emph{Transactions of the Association for Computational Linguistics},
  10:\penalty0 539--554, 2022.

\bibitem[Wang et~al.(2019{\natexlab{a}})Wang, Pruksachatkun, Nangia, Singh,
  Michael, Hill, Levy, and Bowman]{Wang2019-gf}
Alex Wang, Yada Pruksachatkun, Nikita Nangia, Amanpreet Singh, Julian Michael,
  Felix Hill, Omer Levy, and Samuel~R Bowman.
\newblock {SuperGLUE}: A stickier benchmark for {General-Purpose} language
  understanding systems.
\newblock In \emph{Advances in Neural Information Processing Systems}, 32, May
  2019{\natexlab{a}}.

\bibitem[Wang et~al.(2019{\natexlab{b}})Wang, Singh, Michael, Hill, Levy, and
  Bowman]{Wang2019-ll}
Alex Wang, Amanpreet Singh, Julian Michael, Felix Hill, Omer Levy, and Samuel~R
  Bowman.
\newblock {GLUE}: A {Multi-Task} benchmark and analysis platform for natural
  language understanding.
\newblock In \emph{International Conference on Learning Representations},
  2019{\natexlab{b}}.

\bibitem[Wang et~al.(2018)Wang, Yu, Guo, Wang, Klinger, Zhang, Chang, Tesauro,
  Zhou, and Jiang]{Wang2018-zz}
Shuohang Wang, Mo~Yu, Xiaoxiao Guo, Zhiguo Wang, Tim Klinger, Wei Zhang, Shiyu
  Chang, Gerry Tesauro, Bowen Zhou, and Jing Jiang.
\newblock {R$^{3}$}: Reinforced {Ranker-Reader} for open-domain question
  answering.
\newblock In \emph{Proceedings of the {AAAI} Conference on Artificial
  Intelligence}, volume~32. Association for the Advancement of Artificial
  Intelligence, April 2018.

\bibitem[Wei et~al.(2022)Wei, Wang, Schuurmans, Bosma, Chi, Le, and
  Zhou]{Wei2022-lz}
Jason Wei, Xuezhi Wang, Dale Schuurmans, Maarten Bosma, Ed~Chi, Quoc Le, and
  Denny Zhou.
\newblock Chain of thought prompting elicits reasoning in large language
  models.
\newblock In \emph{Thirty-sixth Conference on Neural Information Processing
  Systems ({NeurIPS} 2022)}, January 2022.

\bibitem[Wiegreffe \& Marasovi{\'c}(2021)Wiegreffe and
  Marasovi{\'c}]{Wiegreffe2021-go}
Sarah Wiegreffe and Ana Marasovi{\'c}.
\newblock Teach me to explain: A review of datasets for explainable {NLP}.
\newblock \emph{arXiv:2102.12060 [cs.CL]}, 2021.

\bibitem[Wolf et~al.(2020)Wolf, Debut, Sanh, Chaumond, Delangue, Moi, Cistac,
  Rault, Louf, Funtowicz, Davison, Shleifer, von Platen, Ma, Jernite, Plu, Xu,
  Le~Scao, Gugger, Drame, Lhoest, and Rush]{Wolf2020-ro}
Thomas Wolf, Lysandre Debut, Victor Sanh, Julien Chaumond, Clement Delangue,
  Anthony Moi, Pierric Cistac, Tim Rault, Remi Louf, Morgan Funtowicz, Joe
  Davison, Sam Shleifer, Patrick von Platen, Clara Ma, Yacine Jernite, Julien
  Plu, Canwen Xu, Teven Le~Scao, Sylvain Gugger, Mariama Drame, Quentin Lhoest,
  and Alexander Rush.
\newblock Transformers: State-of-the-art natural language processing.
\newblock In \emph{Proceedings of the 2020 Conference on Empirical Methods in
  Natural Language Processing: System Demonstrations}, Stroudsburg, PA, USA,
  2020. Association for Computational Linguistics.

\bibitem[Wu et~al.(2023)Wu, Zhang, and Huang]{Wu2023-dg}
Dingjun Wu, Jing Zhang, and Xinmei Huang.
\newblock Chain of thought prompting elicits knowledge augmentation.
\newblock In \emph{Findings of the Association for Computational Linguistics:
  {ACL} 2023}, pp.\  6519--6534. Association for Computational Linguistics,
  July 2023.

\bibitem[Xie et~al.(2020)Xie, Thiem, Martin, Wainwright, Marmorstein, and
  Jansen]{Xie2020-xb}
Zhengnan Xie, Sebastian Thiem, Jaycie Martin, Elizabeth Wainwright, Steven
  Marmorstein, and Peter Jansen.
\newblock {{W}orld{T}ree} v2: A corpus of science-domain structured
  explanations and inference patterns supporting multi-hop inference.
\newblock In \emph{Proceedings of the 12th Language Resources and Evaluation
  Conference}, pp.\  5456--5473, Marseille, France, 2020. European Language
  Resources Association.

\bibitem[Xiong et~al.(2021)Xiong, Li, Iyer, Du, Lewis, Wang, Mehdad, Yih,
  Riedel, Kiela, and Oguz]{Xiong2021-ex}
Wenhan Xiong, Xiang Li, Srini Iyer, Jingfei Du, Patrick Lewis, William~Yang
  Wang, Yashar Mehdad, Scott Yih, Sebastian Riedel, Douwe Kiela, and Barlas
  Oguz.
\newblock Answering complex {Open-Domain} questions with {Multi-Hop} dense
  retrieval.
\newblock In \emph{International Conference on Learning Representations}, 2021.

\bibitem[Xu et~al.(2021)Xu, Zhu, Wang, Sun, Cheng, Liu, Gao, He, Zeng, and
  Huang]{Xu2021-ol}
Yichong Xu, Chenguang Zhu, Shuohang Wang, Siqi Sun, Hao Cheng, Xiaodong Liu,
  Jianfeng Gao, Pengcheng He, Michael Zeng, and Xuedong Huang.
\newblock Human parity on {CommonsenseQA}: Augmenting {Self-Attention} with
  external attention.
\newblock \emph{arXiv preprint arXiv: 2112.03254}, December 2021.

\bibitem[Xu et~al.(2022)Xu, Zhu, Wang, Sun, Cheng, Liu, Gao, He, Zeng, and
  Huang]{ijcai2022p383}
Yichong Xu, Chenguang Zhu, Shuohang Wang, Siqi Sun, Hao Cheng, Xiaodong Liu,
  Jianfeng Gao, Pengcheng He, Michael Zeng, and Xuedong Huang.
\newblock Human parity on commonsenseqa: Augmenting self-attention with
  external attention.
\newblock In \emph{Proceedings of the Thirty-First International Joint
  Conference on Artificial Intelligence, {IJCAI-22}}, pp.\  2762--2768, 2022.

\bibitem[Yang et~al.(2018)Yang, Qi, Zhang, Bengio, Cohen, Salakhutdinov, and
  Manning]{Yang2018-xq}
Zhilin Yang, Peng Qi, Saizheng Zhang, Yoshua Bengio, William Cohen, Ruslan
  Salakhutdinov, and Christopher~D Manning.
\newblock {HotpotQA}: A dataset for diverse, explainable multi-hop question
  answering.
\newblock In \emph{Proceedings of the 2018 Conference on Empirical Methods in
  Natural Language Processing}, pp.\  2369--2380. Association for Computational
  Linguistics, 2018.

\bibitem[Yu et~al.(2022)Yu, Zhu, Zhang, Wang, Zhang, Fang, and
  Jiang]{Yu2022-be}
Wenhao Yu, Chenguang Zhu, Zhihan Zhang, Shuohang Wang, Zhuosheng Zhang, Yuwei
  Fang, and Meng Jiang.
\newblock Retrieval augmentation for commonsense reasoning: A unified approach.
\newblock In \emph{Proceedings of the 2022 Conference on Empirical Methods in
  Natural Language Processing}, pp.\  4364--4377. Association for Computational
  Linguistics, October 2022.

\bibitem[Yuan et~al.(2021)Yuan, Neubig, and Liu]{Yuan2021-dm}
Weizhe Yuan, Graham Neubig, and Pengfei Liu.
\newblock Bartscore: Evaluating generated text as text generation.
\newblock \emph{Advances in Neural Information Processing Systems},
  34:\penalty0 27263--27277, 2021.

\end{thebibliography}
